\documentclass[letterpaper, 10 pt, conference]{ieeeconf}  

\IEEEoverridecommandlockouts                              

\usepackage{graphics} 
\usepackage{epsfig} 
\usepackage{amsmath} 
\usepackage{amssymb}  
\usepackage{cite}
\usepackage{color}
\usepackage{balance}
\usepackage{booktabs}
\usepackage{colortbl}

\usepackage{times}

\usepackage[bookmarks=true]{hyperref}
\usepackage{amsfonts}
\usepackage{algorithmic}
\usepackage{algorithm}
\usepackage{array}
\usepackage{textcomp}
\usepackage{stfloats}
\usepackage{url}
\usepackage{verbatim}
\usepackage{mathtools}
\usepackage{amssymb}
\usepackage{multicol}
\usepackage{tikz}
\usepackage{xcolor}
\usepackage{tabularx}
\usepackage{tabularray}
\usepackage{threeparttable}
\usepackage{multirow}

\makeatletter
\let\MYcaption\@makecaption
\makeatother

\usepackage[font=footnotesize]{subcaption}

\makeatletter
\let\@makecaption\MYcaption
\makeatother

\usepackage{subcaption}

\title{\LARGE \bf
Risk-aware Integrated Task and Motion Planning for Versatile Snake Robots under Localization Failures}

\author{Ashkan Jasour$^{*1}$, Guglielmo Daddi$^{*2}$, Masafumi Endo$^{*3}$, Tiago S. Vaquero$^{*1}$, Michael Paton$^1$,\\ Marlin P. Strub$^1$, Sabrina Corpino$^2$, Michel Ingham$^1$,  Masahiro Ono$^{1\dagger}$, Rohan Thakker$^{1\dagger}$
\thanks{$^{*}$Equal contributors.
$^{\dagger}$Equal advisors.}
\thanks{The research was carried out at the Jet Propulsion Laboratory, California Institute of Technology, under a contract with the National Aeronautics and Space Administration (80NM0018D0004). JSPS KAKENHI Grant Number JP22J22731 partially supported this work.}
\thanks{(\emph{Corresponding author: Rohan Thakker.} rohan.a.thakker@jpl.nasa.gov)}
\thanks{$^{1}$Jet Propulsion Laboratory, California Institute of Technology, 4800 Oak Grove Drive, Pasadena, CA 91109, United States}
\thanks{$^{2}$Politecnico di Torino, Corso Duca degli Abruzzi 24, TO 10129, Italy}%
\thanks{$^{3}$Space Robotics Group, Department of Mechanical Engineering, Keio University, Yokohama 223-8522, Japan}
}

\begin{document}

\maketitle
\thispagestyle{empty}
\pagestyle{empty}

\begin{abstract}
Snake robots enable mobility through extreme terrains and confined environments in terrestrial and space applications.
However, robust perception and localization for snake robots remain an open challenge due to the proximity of the sensor payload to the ground coupled with a limited field of view.
To address this issue, we propose Blind-motion with Intermittently Scheduled Scans (BLISS) which combines proprioception-only mobility with intermittent scans to be resilient against both localization failures and collision risks.
BLISS is formulated as an integrated task and motion planning (TAMP) problem that leads to a chance-constrained hybrid partially observable Markov decision process (CC-HPOMDP), known to be computationally intractable due to the curse of history.
Our novelty lies in reformulating CC-HPOMDP as a tractable, convex mixed integer linear program.
This allows us to solve BLISS-TAMP significantly faster and jointly derive optimal task-motion plans.
Simulations and hardware experiments on the EELS snake robot show our method achieves over an order of magnitude computational improvement compared to state-of-the-art POMDP planners and $>$ 50\% better navigation time optimality versus classical two-stage planners.
\end{abstract}

\section{Introduction}

\begin{figure}[t!]
    \begin{subfigure}[t]{\linewidth}
        \centering
        \includegraphics[width=0.95\linewidth]{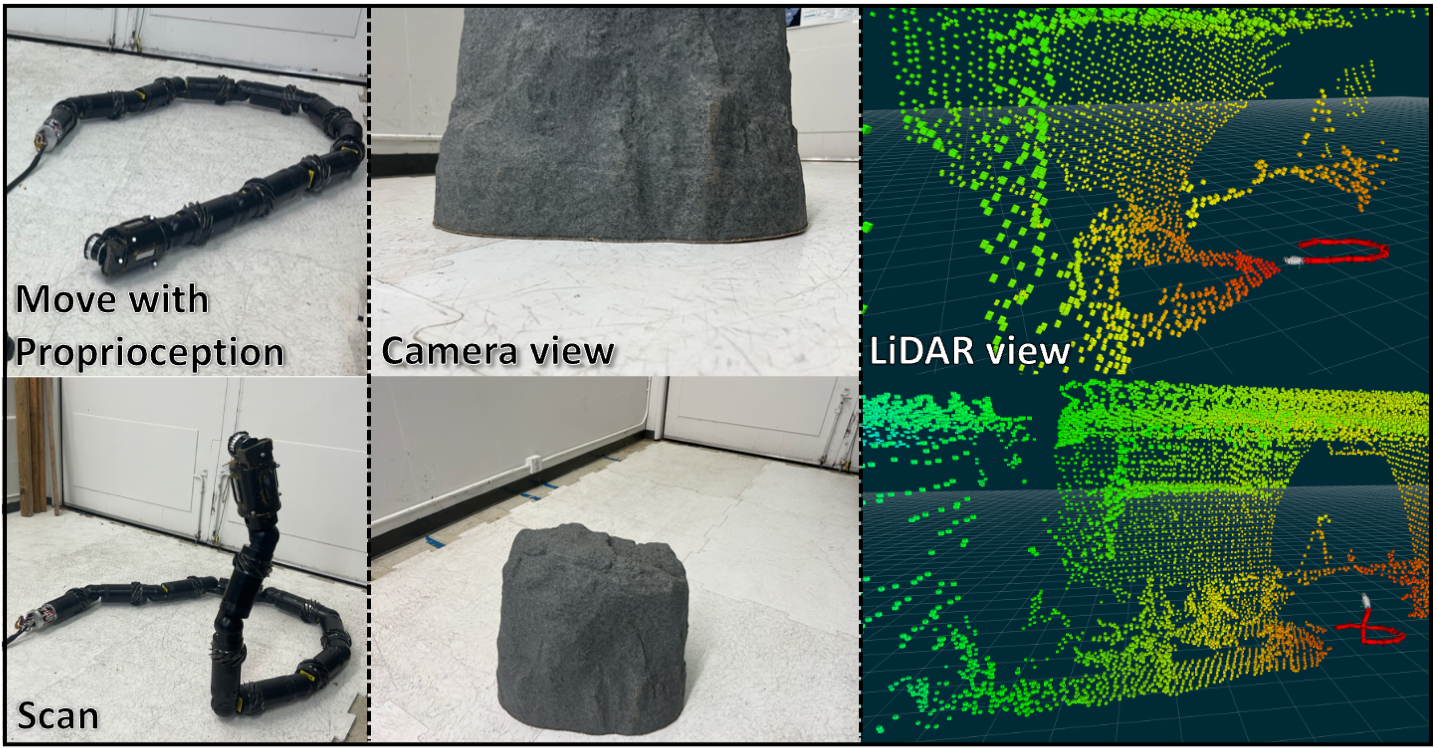}
        \subcaption{Proprioception enables blind movement while scanning collects sensor data. Ground-level sensors increase localization errors. Scanning improves localization and obstacle detection by extending EELS' field of view.}
        \label{fig:move-scan}
    \end{subfigure}
    \\
    \begin{subfigure}[t]{\linewidth}
        \centering
        \includegraphics[width=0.8\linewidth]{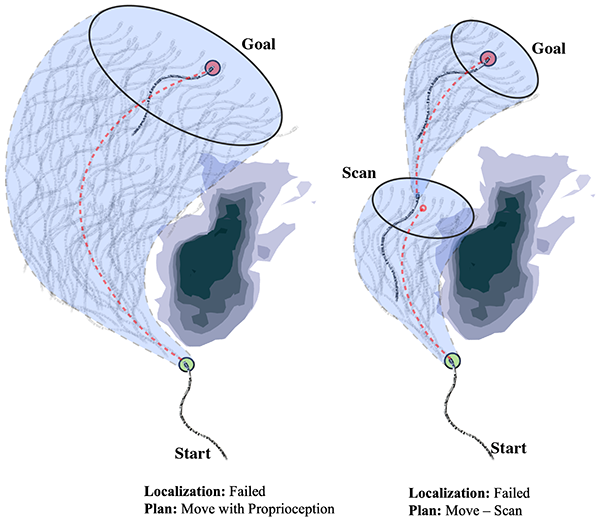}
        \subcaption{Tradeoff between longer paths with higher obstacle clearance but less frequent scans (left) and shorter paths with lower obstacle clearance but more frequent scans (right) to manage collision risk.}
        \label{fig:move-scan-scenario}
    \end{subfigure}
    \caption{Blind-motion with Intermittently Scheduled Scans (BLISS).}
    \label{fig:combined-move-scan}
    \vspace{-5mm}
\end{figure}

Snake robots have versatile mobility for exploring extreme environments due to their reconfigurable body structure.
These robots are versatile for both terrestrial and space missions, from search and rescue in confined spaces to exploring ice worlds like Enceladus or Europa, navigating diverse terrains from vast surfaces to deep crevasses.
However, snake robots struggle with the proximity of their sensor payload to the ground, which increases the complexity of simultaneous localization and mapping.
In featureless or visually challenging environments (\emph{e.g.}, with dust, fog, or snow), this low sensor position limits visible features and accelerates their motion in image space, raising the risk of localization failures.
A restricted view from low-mounted LiDAR also increases the risk of front-end failures in algorithms such as iterative closest point~\cite{talbot2023principled}.
The rapid motion of features in the robot's view further induces sensor data uncertainty and distortion.
Maintaining an upright scan posture, with the robot's front raised, offers a solution, but also consumes energy and limits locomotion capability (Fig.~\ref{fig:move-scan}).
To ensure safe and efficient mobility under uncertainty, snake robots require simultaneous decisions for \emph{where to move} and \emph{when to scan} to reveal unexplored regions.

To this end, we propose Blind-motion with Intermittently Scheduled Scans (BLISS) to achieve resiliency to localization failures by avoiding exteroceptive feedback during motion while leveraging periodic scanning behaviors for localization updates.
As illustrated in Fig. \ref{fig:move-scan-scenario}, BLISS involves solving an integrated task and motion planning (TAMP) problem to balance the tradeoff between longer paths with higher clearance from obstacles but less frequent scans, and shorter paths with lower clearance but more frequent scans.
The challenge is that BLISS-TAMP inherently leads to computationally intractable decision-making, formulated as a chance-constrained hybrid partially observable Markov decision process (CC-HPOMDP).
Thereby, we introduce a novel reformulation as a tractable, convex mixed integer linear program (MILP).
This gives significant computational efficiency than exactly solving CC-HPOMDP and provides better task and motion plans due to its coupled optimization.
Though we focus on snake robots, our solution is applicable to any mobility system capable of proprioceptive locomotion and information-seeking actions with linear dynamics.

We conduct experimental verifications in both simulated and real-world environments.
Simulation results show our approach achieves plans up to 2x more optimal than decoupled task and motion planning, and up to 3x more optimal than a POMDP solver without significant computation. 
Hardware experiments on the EELS snake robot~\cite{thakker2023EELS} further show that the planner can run onboard in a receding horizon fashion, outperforming baseline approaches.

\section{Related Work}
\label{sec:related}

Traditionally, TAMP has been addressed \textit{without considering uncertainty, with decoupled task and motion planning problems}, \emph{i.e.}, solved as distinct sub-problems to combine, usually tasks first then motions.
Examples of this approach are found in the robotic navigation and object manipulation with single and multiple robots \cite{lagriffoul2016combining,dantam2016incremental, he2015towards, akbari2015task, srivastava2014combined, yunl2020petlon, rodriguez2016combining}, and in robotic exploration \cite{fernandezgonzalez2018}. 
Existing work has also explored interleaving or simultaneously addressing task and motion planning without uncertainty \cite{diab2019pmk,cambon2009hybrid,garrett2015ffrob,akbari2015reasoning,akbari2016task,akbari2019knowledge}. 
For instance, \cite{akbari2019combined} combines heuristic-based task planner with geometric reasoners for manipulation planning in table-top object manipulation problems for bi-manual robots. 
In this line of work, motion planning serves as a central part of the action selection process in task planning, \emph{e.g.}, for computing task feasibility at a particular state. 

One of the central themes in our work is addressing \textit{uncertainty} in TAMP. 
Here, an agent plans using incomplete information or probabilistic models of the environment. 
Most existing work on this front interleaves task and motion planning, where the uncertainty is incorporated in task planning with discrete actions and states. 
However, even with discrete spaces, the problem quickly becomes intractable as the search tree branches in both actions and observations. 
In \cite{nouman2017experimental, akbari2020contigent}, conditional or contingent task planning is the core process with geometric evaluation used as low-level feasibility checks for task selection. 
\cite{mochizukiKT22} uses the same principle but employs a probabilistic task planning approach.

We address \textit{TAMP under uncertainty} in the form of chance-constrained TAMP (CC-TAMP), where the system considers constraints on the probability of failure to execute a control plan.
Directly related work includes \cite{santana2016hybrid, huang2018hybrid}, which solves a CC-HPOMDP formulation for autonomous vehicles and Mars rovers in a hierarchical fashion.   
Also, \cite{ayton2019astro, ayton2019aaai} study adaptive sampling in robotic ocean exploration to track a science phenomenon modeled as a Gaussian process. 
We address the CC-TAMP under uncertainty by simultaneously addressing task and motion planning, rather than decoupling them.
Our formulations consider continuous state spaces and both discrete and continuous action spaces that significantly increase the search space and computational complexity.

\begin{table}[t]
    \centering
    \caption{Representative related work landscape}
    \begin{tabular}{l|cccc}
    \hline
    \hline
    \multirow{2}{*}{Work} & Integrated & Under & Chance & Compute \\
    & TAMP & Uncertainty & Constraints & Efficiency \\
    \hline
    \cite{lagriffoul2016combining,dantam2016incremental, he2015towards, akbari2015task, srivastava2014combined, yunl2020petlon, rodriguez2016combining, fernandezgonzalez2018}
            &              &               &    &      \\
    \hline
    \cite{diab2019pmk, cambon2009hybrid, garrett2015ffrob, akbari2015reasoning, akbari2016task, akbari2019knowledge, akbari2019combined}
            & \checkmark    &               &      &    \\
    \hline
    \cite{nouman2017experimental, akbari2020contigent, mochizukiKT22}
            &  \checkmark   &  \checkmark   &       &   \\
    \hline
    \cite{santana2016hybrid, ayton2019astro, ayton2019aaai}
            &               &  \checkmark   & \checkmark  &\\
    \hline
    This work
            & \checkmark    &  \checkmark   & \checkmark  & \checkmark\\
    \hline
    \hline
    \end{tabular}    
    \label{tab:relatedwork}
    \vspace{-4mm}
\end{table}

Given our focus on risk-awareness, we address the computationally challenging \emph{risk-aware motion planning problem} in uncertain environments.
Sampling-based methods, such as those by \cite{janson2017monte,blackmore2010}, estimate safety constraint satisfaction probabilities by sampling uncertainties. 
\cite{blackmore2011chance} reformulates the problem as a MILP for polygon obstacles and linear Gaussian systems, while \cite{blackmore2009convex} proposes a linear program assuming convex safe regions.
By adapting traditional algorithms, \cite{luders2010chance} develops a chance-constrained RRT for motion planning under uncertainty. 
For nonlinear constraints, \cite{lew2020chance,dai2019chance} use linearized models and Gaussian approximations, whereas \cite{han2022non, jasour2021convex} propose nonlinear algorithms for non-Gaussian uncertainties, considering higher-order moments.
Our proposed method builds upon existing work in risk-aware motion planning, while providing direct uncertainty incorporation rather than uncertainty sampling.

Our contributions are summarized as follows:\\
i) We introduce BLISS for systems with limited simultaneous motion and localization to achieve blind mobility and schedule intermittent scans for localization and obstacle avoidance.\\
ii) We formulate BLISS as a CC-HPOMDP to integrate task and motion planning under uncertainties, which has not been achieved by existing methods (Table\ref{tab:relatedwork}).\\
iii) We provide an efficient convex deterministic optimization that ensures safety guarantees by satisfying probabilistic constraints without relying on uncertainty samples.

\section{Problem Formulation}

\subsection{BLISS-TAMP for Long-range Navigation}
Robust perception and localization for snake robots remain an open problem.
The use of snake robot reconfigurability for scanning behaviors was first explored in \cite{ponte2014movescan}, where the authors focused on hardware and gait design with limited onboard autonomy.
We leverage the EELS robot's scanning capability for long-range localization and obstacle detection, combined with blind proprioception-only mobility for navigation, to be resilient against localization failures when exteroceptive sensors are near the ground.

We propose a long-range, risk-aware task and motion planner for the BLISS in extreme terrains.
This planner views the robot at an abstract level to approximate it as a point mass with linear dynamics.
It creates a low-resolution 2.5-D motion path with scheduled scans that meet a predefined risk tolerance $\Delta_c$.
A short-range planner and controllers execute the remaining process by accounting for the robot's high degrees of freedom (DoF) and non-linear dynamics~\cite{thakker2023EELS}.
In summary, Fig. \ref{fig:long-vs-short-range-planner-architecture} illustrates this layered autonomy from long-range planning to high DoF control.
Details of other autonomy components can be found in \cite{thakker2024to, vaquero2024eels}.

\subsection{CC-HPOMDP for BLISS-TAMP under Uncertainty}
\begin{figure}[t]
    \centering
    \includegraphics[width=0.7\linewidth]{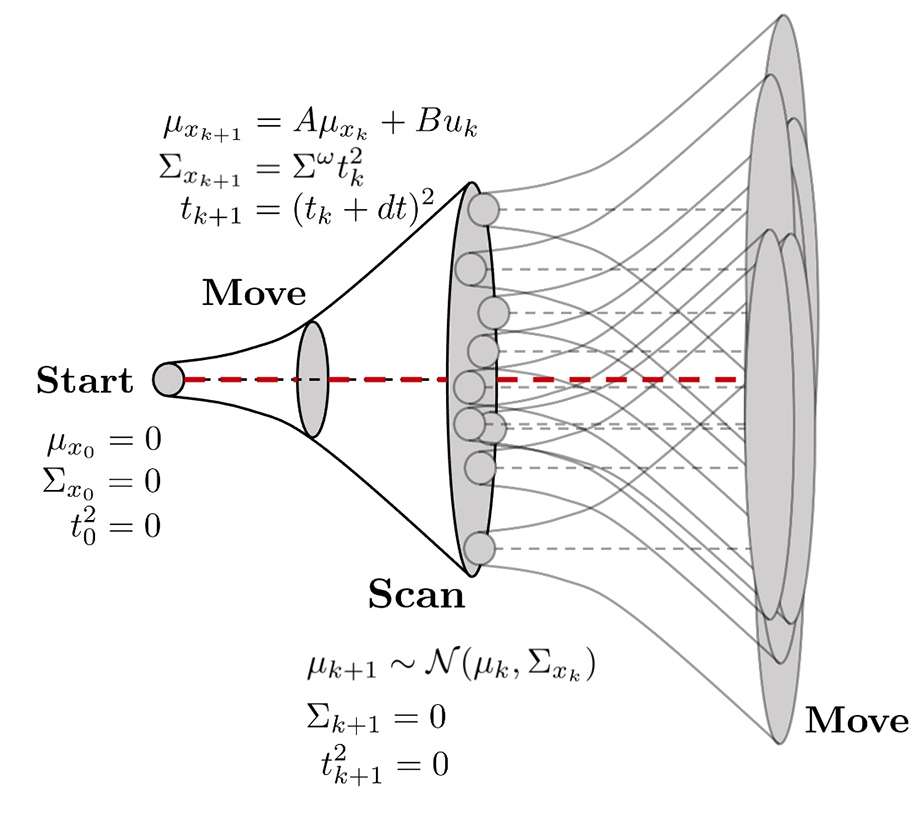}
    \vspace{-2mm}
    \caption{BLISS-TAMP introduces hyperbeliefs. After initial scanning, the state's mean depends on the observation. This transforms belief into a distribution over distributions for post-scan state prediction.}
    \label{fig:hyperbelief}
    \vspace{-4mm}
\end{figure}

BLISS-TAMP can be interpreted as an agent reaching a goal through actions and observations subject to safety constraints.
Due to sensor noise and imperfect models, both state and observations are expected to be random variables.
The agent's action space is both discrete (move vs scan) and continuous (where to move).
This problem structure aligns with CC-TAMP and can be cast as a CC-HPOMDP due to these uncertainties.
We represent uncertain state as a continuous random variable $X : \mathbb{R}^{N_X} \to \mathbb{R}$, continuous control input as $u \in \mathbb{R}^{N_u}$, and
discrete actions as $a \in \mathcal{A} = \{1, 2, ... , N_a\}$.
Action and time dependence are expressed as $\prescript{a}{}{v}_k^i$ where the $i$ denotes the component of multi-dimensional variable $v$, $k$ the time step, and $a$ the associated action.
This problem can be formulated as:
\begin{gather}
    CC-HPOMDP = \langle S, A, T, \Omega, R, \gamma, \mathcal{Z}, \mathcal{C}, \Delta_c \rangle. \label{eq:CC-HPOMDP}
\end{gather}
$S$ is the state space, $A$ the action space, $T = p(s_{k+1}|s_k,a_k, u_k)$ the transition function, $\Omega$ the observation space, $\mathcal{Z} = p(o_{k+1}|s_{k+1}, a_k, u_k)$ the observation function, $R$ the reward function, $\gamma$ the discount factor, $\mathcal{C}$ set of state constraints, $\Delta_c \subset [0,1]$ risk bound \cite{cassandra1994acting}.
CC-HPOMDP planning occurs in belief space - probability distribution over states, and requires tracking of probability distributions of probability distributions, \emph{i.e.,} \textit{hyperbeliefs} (Fig.~\ref{fig:hyperbelief}).

\subsection{Mathematical Optimization Program for CC-HPOMDP}
We cast CC-HPOMDP as a mathematical optimization program with cost minimization objective over $N$ time steps:
\begin{subequations}
\begin{gather}
    \min_{X_{0:N},u_{0:N},a_{0:N-1}} \mathbb{E}\left[\sum_{k=0}^{N-1} c_kX_k \right], \label{eq:cost}\\
    X_{k+1} = \prescript{a_k}{}{A}X_k + \prescript{a_k}{}{B}u_k + \prescript{a_k}{}{\omega}_k + \prescript{a_k}{}{C} \label{eq:noisy-transition},\\
    X_0 = B_0(x), \quad {\mathbb{E}[X_N] = x_{\text{goal}}}, \quad Tr(\mathbb{V}[X_N]) \leq \epsilon^{\Sigma}, \label{eq:initial-terminal}\\
    \mathrm{Pr}(X_k\in X^{\text{obs}}) < \Delta_c. \label{eq:chance-constrains}
\end{gather}
\end{subequations}
\noindent (\ref{eq:cost}) denotes the cost function with the given cost coefficient vector $c_k$. (\ref{eq:noisy-transition}) represents the system transition dynamics where $\prescript{a_k}{}{A},\prescript{a_k}{}{B}$ and $\prescript{a_k}{}{C}$, $\prescript{a_k}{}{\omega}_k$  are the state transition matrix, control input matrix, bias of the system, and process noise of the system, respectively, under discrete action $a$ at time $k$. 
(\ref{eq:initial-terminal}) defines the initial and terminal conditions: $B_0(x)$ is the initial belief, $x_{\text{goal}}$ the goal state, and $Tr(\mathbb{V}[X_N])$ the trace of the state's covariance that models the state uncertainty size at the planning horizon's end where $\epsilon^{\Sigma}$ denotes the maximum allowable uncertainty. 
(\ref{eq:chance-constrains}) represents the probabilistic safety constraints relative to obstacle sets $X^{\text{obs}}$ with $\Delta_c$ as the given acceptable risk tolerance.

\textit{Remark 1: 
In the presence of multiple obstacles, one can apply separate chance constraints, as specified in \eqref{eq:chance-constrains}, for each obstacle with different risk levels. To meet the total risk level of $\Delta_c$, one can allocate the risk uniformly, assigning $\Delta_c/N_o$ to each obstacle where $N_o$ is the number of obstacles, or utilize different risk allocation methods such as \cite{risk_allo1}.} 

\textit{Remark 2: The states $X_k$ are not observable because $X_0$ is a random variable, and at each time $k$, the transition dynamics introduce noise $\omega_k$. Hence, this planning problem needs to be interpreted as a CC-HPOMDP framework.}

\begin{figure}[t]
    \centering
    \includegraphics[width=0.99\linewidth]{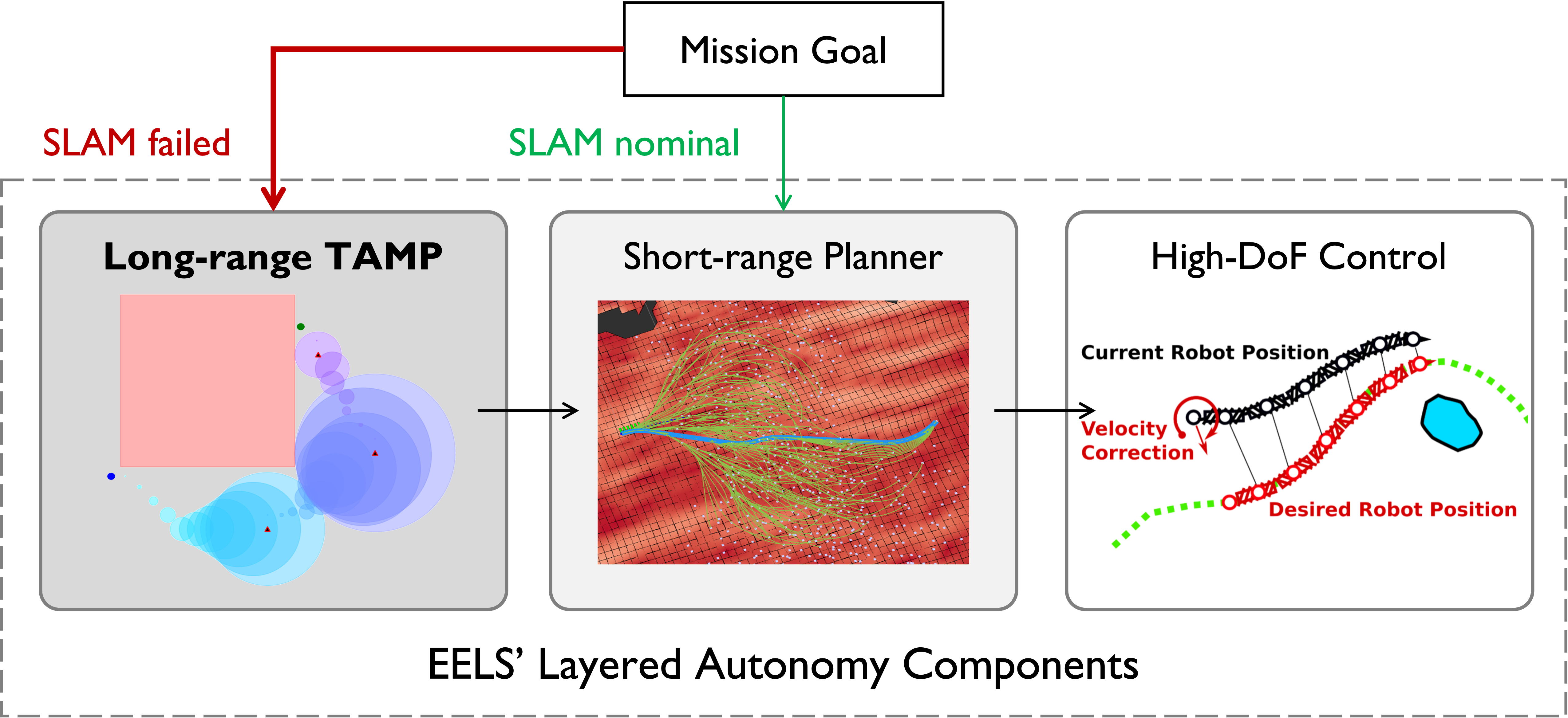}
    \caption{EELS' layered autonomy architecture: from Long-range TAMP (this work), Short-range planner, to High-DoF controllers.}
    \label{fig:long-vs-short-range-planner-architecture}
    \vspace{-4mm}
\end{figure}

\subsection{Belief MDP Formulation for BLISS-TAMP}
\label{sec:bmdp-move-scan}
BLISS-TAMP owns two discrete actions: $Move$ and $Scan$, denoted by prescripts $M$ and $S$, respectively.
$Move$ changes the agent's state using proprioception and without observations, while $Scan$ re-localizes the agent in the map.
The state vector $X$ comprises the agent's Cartesian coordinates $x$ and $y$.
The control input $u=[v_x, v_y]^T$ is the velocity in $x,y$ dimensions.
We cast the BLISS-TAMP's CC-HPOMDP formulation into belief space, \emph{i.e.}, belief MDP, assuming the agent's state uncertainty is normally distributed $X \sim \mathcal{N}(\mu_X, \Sigma_X)$, and describing state transition dynamics as a function of this belief state.
If observations remained random variables, the planning problem would be in hyperbelief space, which is typically intractable~\cite{hauskrecht2000value}.
Thus, we assume the agent's post-scan state is its pre-scan belief maximum likelihood estimate.
Specifically, after scanning, the agent's true position is assumed to match the expected value of its pre-scan state probability distribution.
While this simplification breaks plan optimality and safety guarantees~\cite{cunningham2015mpdm, leung2006planning}, good results can be achieved by combining it with \textit{model predictive control}, which involves re-planning after each observation during execution.
We assume an identity transformation between observations and states.

\textit{Remark 3: State covariance depends on the action-selection decision variable as scanning re-localizes the agent in the map. 
This is a non-linearity that can be removed by assuming a state covariance growth that is quadratic over time since the last scanning behavior was performed.}

We augment the state vector with the time since the last information-gaining action was taken $t$ and $t^2$.
The transition matrices can thus be written as:
\begin{subequations}
    \begin{gather}
    N_X = 4, \quad N_u = 4, \quad N_a = 2, \nonumber\\
    \mathcal{A} = \{Move = M, Scan = S\}, \nonumber\\
    X = \begin{bmatrix}
        x & y & t^2 & t 
    \end{bmatrix}^T, \\
   u = \begin{bmatrix}
        v_x & v_y & 0 & 0 
    \end{bmatrix}^T, \\
    \prescript{M}{}{A} = 
    \begin{bmatrix}
    1 & 0 & 0 & 0 \\
    0 & 1 & 0 & 0 \\ 
    0 & 0 & 1 & 2\Delta t \\
    0 & 0 & 0 & 1
    \end{bmatrix}, 
    \prescript{M}{}{B} = \begin{bmatrix}
        \Delta t & 0 & 0 & 0 \\
        0 & \Delta t & 0 & 0 \\ 
        0 & 0 & 0 & 0 \\
        0 & 0 & 0 & 0
    \end{bmatrix} \label{eq:cc-tamp-a-b-mat1} \\ 
    \prescript{S}{}{A} = \begin{bmatrix}
    1 & 0 & 0 & 0 \\
    0 & 1 & 0 & 0 \\ 
    0 & 0 & 0 & 0 \\
    0 & 0 & 0 & 0
    \end{bmatrix}, \prescript{S}{}{B} = 0_{4x4} \label{eq:cc-tamp-a-b-mat2} \\
    \prescript{M}{}{\omega} =  \mathcal{N}(0, \Sigma^\omega), \prescript{S}{}{\omega} =  0 \\
    \prescript{M}{}{C} = \begin{bmatrix}
        0 & 0 & (\Delta t)^2 & \Delta t
    \end{bmatrix}^T,
    \prescript{S}{}{C} = \begin{bmatrix}
        0 & 0 & 1 & 1
    \end{bmatrix}^T \label{eq:cc-tamp-c-mat}
\end{gather}
\label{eq:cc-tamp-mat}
\end{subequations}
\noindent Matrices $\prescript{M}{}{A}$, $\prescript{M}{}{B}$ and $\prescript{M}{}{C}$ show how the agent's coordinates are modified by the control input.
Time advances when the agent moves.
The scan action maintains the agent's coordinates and resets time $t$ to $1$. 
State variables $x$ and $y$ represent the robot's center of mass.
As BLISS-TAMP is a long-range planner (Fig. \ref{fig:long-vs-short-range-planner-architecture}), we can treat the robot as a point mass and omit the snake robot dynamics. 
The resulting path is tracked by a short-range planner that explicitly considers the robot's high DoF and non-linear dynamics.
The time state variables are propagated following this expression:
\begin{gather}
    t_{k+1} = t_k + \Delta t,\\
    t^2_{k+1} = (t_k + \Delta t)^2 = t_k^2 + 2 t_k \Delta t + \Delta t^2.
\end{gather} 
Belief transition dynamics can be written as:
\begin{align}
\mu_{X_{k+1}} &= 
\prescript{a_k}{}{A} \mu_{X_k} + \prescript{a_k}{}{B} u_k + \prescript{a_k}{}{\omega_k} + \prescript{a_k}{}{C},\\
\Sigma_{X_{k+1}} &= 
\prescript{a_k}{}{A} \Sigma_{X_{k}} \prescript{a_k}{}{A^T} + \prescript{a^i}{}{\Sigma_{\omega_k}}. \label{eq:cov-transition-non-milp}
\end{align}
Transition dynamics can be combined into a single equation by introducing $N_a$ additional binary selection variables for control discrete actions $\prescript{a^i}{}{y}_k \in \{0,1\}$ as follows:
\begin{align}
    \mu_{X_{k+1}} &=  
    \sum_{i = 1}^{N_a} \prescript{a^i_k}{}{y_k} (\prescript{a^i_k}{}{A} \mu_{X_k} + \prescript{a^i_k}{}{B} u_k + \prescript{a^i_k}{}{\omega_k} + \prescript{a_k^i}{}{C}),\label{eq:unc_prop_1}\\
    \Sigma_{X_{k+1}} &=  \sum_{i = 1}^{N_a}
    \prescript{a^i_k}{}{y_k} (\prescript{a^i_k}{}{A} \Sigma_{X_{k}} \prescript{a^i_k}{}{A^T} + \prescript{a^i_k}{}{\Sigma_{\omega_k}}), \label{eq:unc_prop_2}
\end{align}
\begin{equation}
    \sum_{i = 1}^{N_a} \prescript{a^i_k}{}{y_k} = 1, \label{eq:unc_prop}
\end{equation}
where $\prescript{a^i}{}{y}_k =1$ if action $a^i$ is executed at time $k$. 
Constraints in \eqref{eq:unc_prop_1} and \eqref{eq:unc_prop_2} are bi-linear constraints due to the product of optimization variables, e.g., $ \prescript{a^i_k}{}{y_k} u_k$, $ \prescript{a^i_k}{}{y_k} {\mu_X}_k$. 
Any optimization variable $h_k \in [h_{\min}, h_{\max}]$ multiplied by $\prescript{a^i_k}{}{y_k}$ is linearized using a \emph{slack variable} $\prescript{a^i}{}{r^{h}_k} = h_k \prescript{a^i}{}{y_k}$ with additional constraints:
\begin{subequations}
\begin{gather}
    {h}_{\min}\prescript{a^i}{}{y_k} \le r^{h}_k \le {h}_{\max}\prescript{a^i}{}{y_k}, \label{eq:slack_1}\\
    h_k - {h}_{\max} (1-\prescript{a^i}{}{y_k}) \le r^{h}_k \le h_k - {h}_{\min} (1-\prescript{a^i}{}{y_k}).
    \label{eq:slack_2}
\end{gather}
\end{subequations}
Linearization is applied to $\prescript{a^i}{}{r}^{\mu_X}_k = \mu_{X_k} \prescript{a^i}{}{y_k}$ and  $\prescript{a^i}{}{r}^{u}_k = u_{k} \prescript{a^i}{}{y_k}$. 
$\Sigma^X$ does not require a slack variable, \emph{as the agent's state covariance follows the quadratic time growth model} $\Sigma_{k+1}^X = \Sigma^\omega t^2_k$.
The state's mean transition constraints can be written as: 
\begin{gather}
    \mu_{X_{k+1}} = \sum_{i=1}^{N_a} \left( \prescript{a^i}{}{A}\prescript{a^i}{}{r}_k^{\mu_X} + \prescript{a^i}{}{B}\prescript{a^i}{}{r}_k^u + \prescript{a^i}{}{C}\right).
    \label{eq:cc-tamp-milp-transition}
\end{gather}

\emph{Chance Constraints:} \eqref{eq:chance-constrains} are applied for obstacle collision avoidance.
The unsafe region $O$ consists of $N_o$ polygonal obstacles. 
With $N^o_e$ edges for the $o^{th}$ unsafe region, these constraints form a disjunctive program \cite{blackmore2011chance,da2019collision}:
\begin{equation}
\label{eq:disjuntive-obstacle}
\bigcup_{i=1}^{N^o_e} \left[H^o_i\mu_{X_k} + \phi^{-1}(\Delta_c) {H^o_i}^T\Sigma_{X_k}H^o_i \geq b^o_i \right].
\end{equation}
$H^o_i$ and $b^o_i$ describe the coefficients of the $i^{th}$ half-plane, with $\phi$ as the standard normal distribution. 
(\ref{eq:disjuntive-obstacle}) applies to each obstacle at every time step.
We write the disjunctive program as a mixed integer program using big-M constraints: 
\begin{subequations}
\begin{gather}
    \bigcap_{i=1}^{N^o_e} \left[ H^o_i\mu_{X_k} + \phi^{-1}(\Delta_c) {H^o_i}^T\Sigma_{X_k}H^o_i \geq \prescript{o}{}{b}_i - M^o(1 - y^{o,i}_k) \right],\\ 
    \sum_{i=1}^{N^o_e}y^{o,i}_k = \prescript{o}{}{N}_e - 1.
    \label{eq:cc-tamp-cc}
\end{gather}
\end{subequations}

\section{Approach: Convex MILP for BLISS-TAMP}
\label{sec:BLISS-TAMP MILP}
We describe how the BLISS-TAMP's belief MDP formulation can be cast as a tractable MILP. 
Building on the formulation introduced in Section \ref{sec:bmdp-move-scan}, we propose an objective function that jointly penalizes scanning and path length.
To minimize the time spent outside of the goal, we introduce binary variables $z^{\text{og}}_k\in\{0,1\}$ that act as a goal indicator function. 
These variables are zero when the agent is in the \emph{goal set}, and 1 otherwise.
The goal set is defined as the set of agent belief states where the state mean is within a threshold $\epsilon^g \in \mathbb{R}$ from the goal point $(x^g,y^g)$, and the trace of state covariance is below $\epsilon^\Sigma$.
Additional constraints on the binary variables complete the problem formulation:

\begin{subequations}
\begin{align}
    \min &\sum_{i=0}^N \left( \prescript{S}{}{c}_k\prescript{S}{}{y}_k + \prescript{M}{}{c}_kz^{\text{og}}_k\right) \label{eq:cc-tamp-milp-opt-obj-start}, \\
    z^{\text{og}}_k &\in \{0,1\}, \\
    a_k^x + a_k^y - \epsilon^{g} &\leq M^gz^{\text{og}}_k, \\
    \epsilon^{g} - a_k^x - a_k^y& \leq M^g(1 - z^{\text{og}}_k), \\
    x_k - x_{\text{goal}} &\leq a_k^x \leq -x_k + x_{\text{goal}}, \\
    y_k - y_{\text{goal}}& \leq a_k^y \leq -y_k + y_{\text{goal}}.
    \label{eq:cc-tamp-milp-opt-obj-end}
\end{align}
\label{eq:cc-tamp-milp-opt-obj}
\end{subequations}
Here, $a_k^x\in\mathbb{R}^+$ and $a_k^y\in\mathbb{R}^+$ are additional variables introduced to linearize $|x - x_{\text{goal}}|$ and $|y-y_{\text{goal}}|$.
Additional constraints are required to prevent corner-cutting \cite{da2019collision}, as MILP formulations lack collision checking between states.
\begin{subequations}
\begin{gather}
    \Sigma_0 t_k^2 - M^c(1-z_k^{\text{ig}}) \leq \Sigma_{k+1} \leq \Sigma_0 t_k^2 + M^c(1-z_k^{\text{ig}}),\label{eq:cc-tamp-cov-prop-1}\\
    \Sigma_k - z_k^{\text{ig}}M^c \leq \Sigma_{k+1} \leq \Sigma_k + z_k^{\text{ig}}M^c,\\
    z_k^{\text{ig}} \in \{0,1\},\ z_k^{\text{og}} + z_k^{\text{ig}} = 1. \label{eq:cc-tamp-cov-prop-2}
\end{gather}
\label{eq:cc-tamp-cov-prop}
\end{subequations}

\textit{Remark 4:
We showed that BLISS can be formulated as a convex MILP by: 1) introduction of slack variables, e.g., $r_k^h$, and 2) treating time (since the last scan) $t$ as a separate system state and describing the state covariance propagation in terms of time $t$. Otherwise, the MILP formulation of BLISS problem will result in a nonconvex optimization due to bi-linear terms, e.g., the product of decision variables.}

In summary, the convex MILP formulation for BLISS-TAMP optimizes over all variables to solve \eqref{eq:cc-tamp-milp-opt-obj} subject to mean state transition dynamics (\ref{eq:cc-tamp-milp-transition}) with matrices (\ref{eq:cc-tamp-a-b-mat1}, \ref{eq:cc-tamp-a-b-mat2}, \ref{eq:cc-tamp-c-mat}); slack variable constraints (\ref{eq:slack_1}) and (\ref{eq:slack_2}) for state variables multiplied by selector variables; covariance propagation \eqref{eq:cc-tamp-cov-prop}; and polygonal obstacle chance constraints (\ref{eq:cc-tamp-cc}).

\section{Experiments}
This section evaluates our MILP planner through simulations and hardware tests on the EELS snake robot \cite{thakker2023EELS}.

\subsection{Metrics of Interest}
We evaluate planners on navigation efficiency and computation efficiency.
\emph{Success Rate} (SR) [\%] indicates the percentage of successful executions to reach the goal position.
\emph{Execution Time} (ET) [s] measures navigation efficiency as the total time from start to goal positions. 
The required times for $Move$ and $Scan$ actions are set as 0.5 s and 100 s respectively.
\emph{Computation Time} (CT) [s] evaluates computation efficiency by total time spent for planning. ET and CT are calculated for successful executions.

\begin{figure}[t]
  \centering
  \begin{minipage}[b]{0.333\linewidth}
    \includegraphics[width=\linewidth]{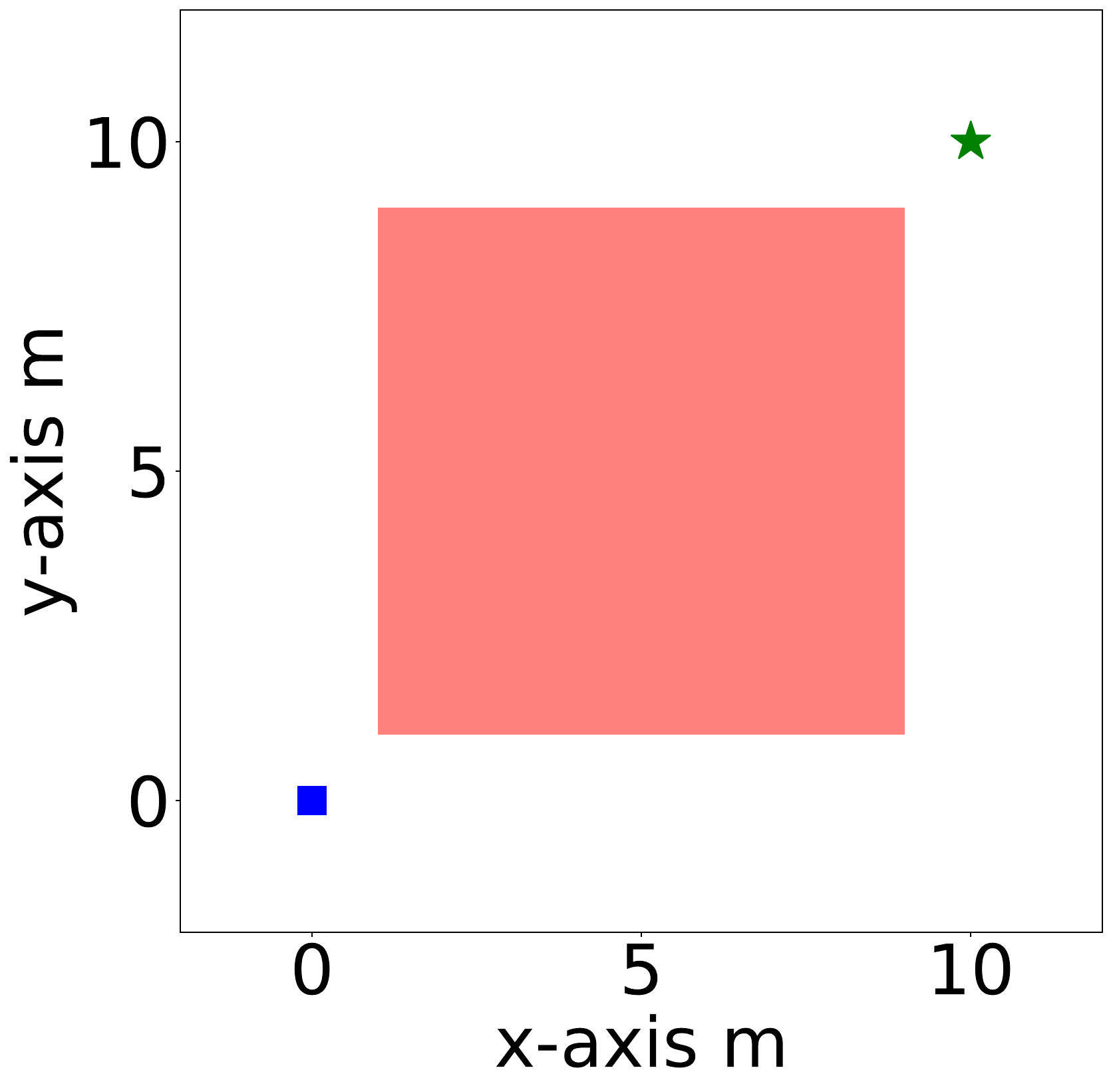}
    \subcaption{Standard map}
  \end{minipage}%
  \begin{minipage}[b]{0.333\linewidth}
    \includegraphics[width=\linewidth]{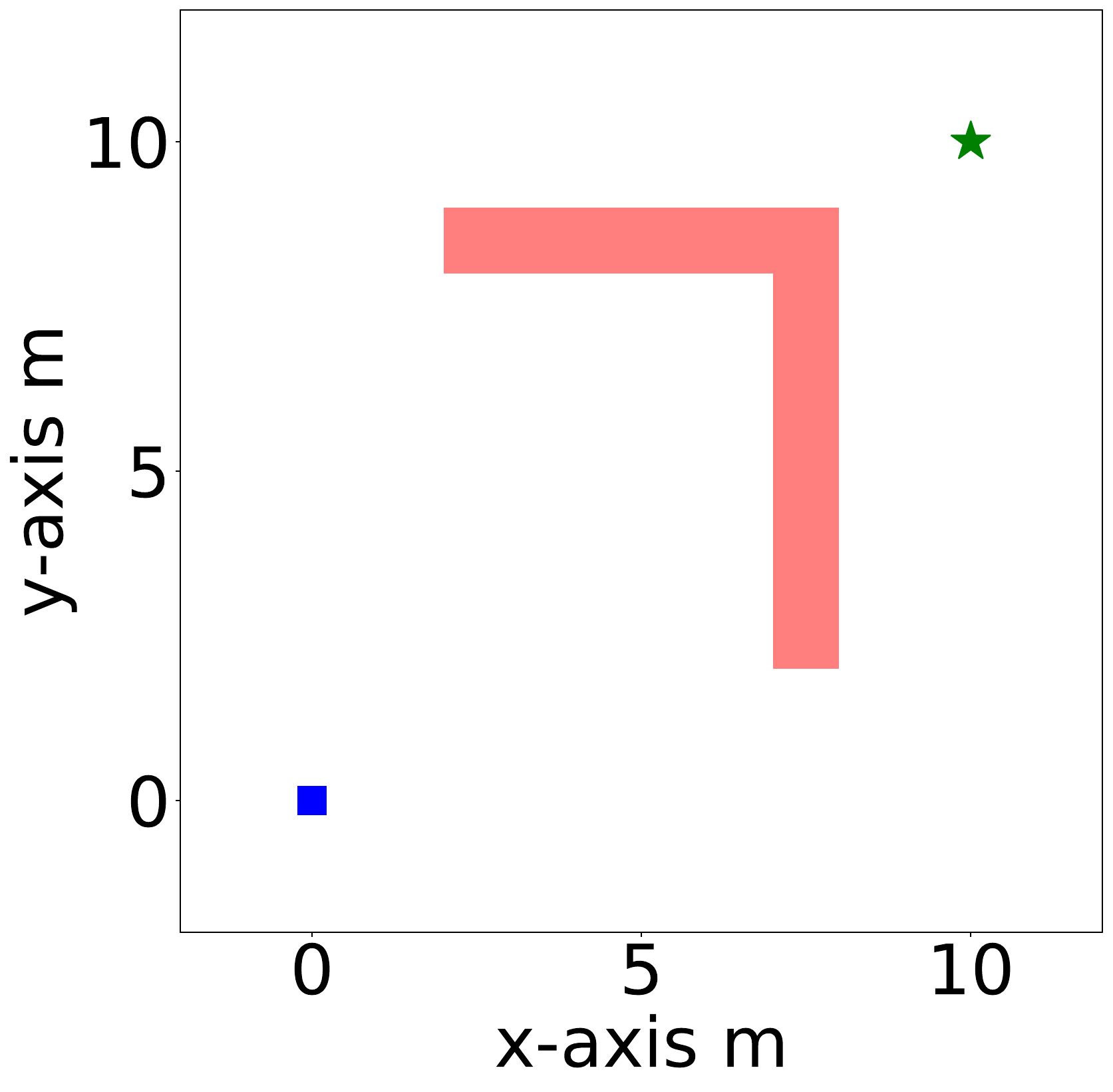}
    \subcaption{Entrapped map}
  \end{minipage}%
  \begin{minipage}[b]{0.333\linewidth}
    \includegraphics[width=\linewidth]{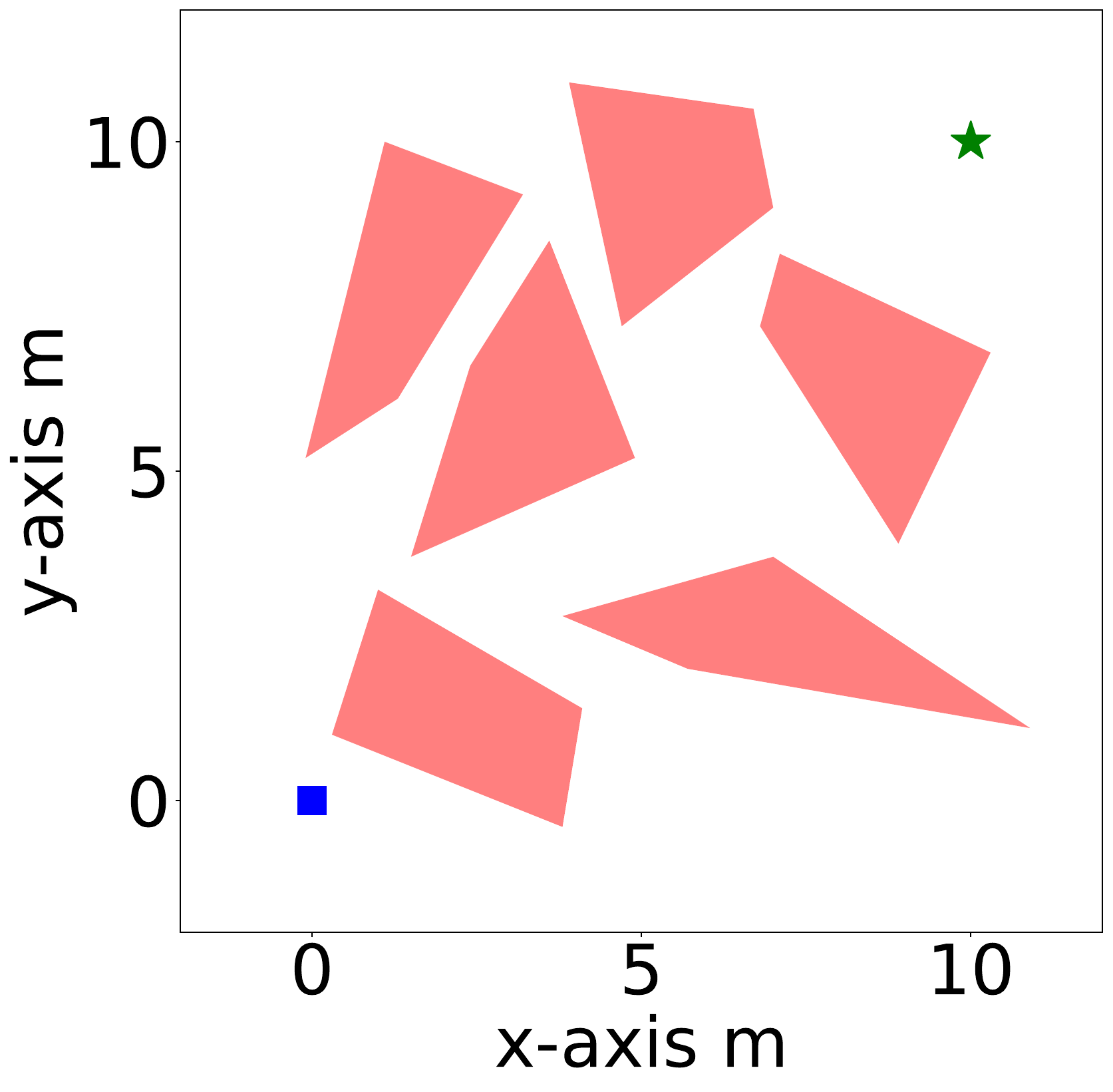}
    \subcaption{Random map}
  \end{minipage}
  \vspace{-4mm}
  \caption{Obstacle scenarios (red) for simulation studies, with start and goal positions, colored in blue and green, respectively.}
  \vspace{-4mm}
\label{fig:maps}
\end{figure}

\subsection{Experimental Setups}
Our approach is compared with two baseline planners:
\emph{Two-stage Planner:} This approach follows the classic decoupling of path planning and task scheduling, which are handled separately. 
A* search plans paths on a 2D gridmap with inflated obstacles and scan scheduling follows path planning. 
Uncertainty is pre-computed at each step; scans are scheduled only when risk constraints are violated. 
The planner repeats path planning and scheduling after each scan.

\emph{POMDP Planner:} This approach solves the CC-HPOMDP problem using MCTS-DPW \cite{couetoux2011continuous}, a state-of-the-art POMDP solver.
This formulation requires action space discretization, so the control input is discretized into 8 equal components with continuous state space.
Yet, this baseline exhibits overly myopic behaviors, failing to reach goals due to the POMDP's limited planning horizon. 
To ensure a fair comparison, we extend it using \cite{kim2019bi} that enables long-horizon planning through belief space roadmaps from goal to start.

We evaluate planners using a simulator that generates robot behavior following the dynamics described in Section \ref{sec:BLISS-TAMP MILP}.
The two-stage and POMDP planners are implemented in Julia, using the Julia POMDP library \cite{egorov2017POMDPs}. 
The MILP planner is implemented in C++ using Google's or-tools \cite{ortools} and Gurobi \cite{gurobi}.
We integrate these multi-language configurations using ROS that connect planning and execution components through a client-service architecture.
Note that all planners run on the same machine (Intel Core i7-1165G7 CPU, Ubuntu) to ensure a fair comparison of CT.

We prepared four obstacle scenarios: Standard (Std), Entrapped (Ent), Narrow (Nar), and Random (Rnd), where Std, Ent, and Rnd are illustrated in Fig. \ref{fig:maps}a, b, and c, respectively.
Std-map assesses basic collision avoidance in open spaces.
Ent-map tests planners against local minima entrapment due to finite horizon planning.
Nar-map examines the trade-off between longer paths with fewer scans and shorter paths requiring more scans.
Rnd-map with six quadrilaterals tests general performance in obstacle-rich scenarios.
In all scenarios, planners search from (0 m, 0 m) to (10 m, 10 m).

\begin{table*}[t!]
    \centering
    \begin{threeparttable}        
    \caption{Quantitative chance-constrained TAMP performance results on three obstacle scenarios}
   \label{tab:quantitative_comparison_three_configs}
    \setlength{\tabcolsep}{3.0pt}
        \begin{tabular}{ll|ccc|ccc|ccc}
        \toprule
        & & \multicolumn{3}{c|}{Std-map} & \multicolumn{3}{c|}{Ent-map} & \multicolumn{3}{c}{Nar-map}\\
        \cmidrule(lr){1-2} \cmidrule(lr){3-5} \cmidrule(lr){6-8} \cmidrule(lr){9-11} 
        \multicolumn{2}{l|}{Planners} & SR$\uparrow$ & ET$\downarrow$ & CT$\downarrow$ & SR$\uparrow$ & ET$\downarrow$ & CT$\downarrow$ & SR$\uparrow$ & ET$\downarrow$ & CT$\downarrow$\\
        \midrule \midrule
        \multicolumn{2}{l|}{Two-stage} & 72 & 278.69 $\pm$ 59.38 & \textbf{4.84} $\pm$ 8.29 & 88 & 166.00 $\pm$ 21.45 & \textbf{4.59} $\pm$ 7.89 & 64 & 412.09 $\pm$ 0.27 & \textbf{4.57} $\pm$ 8.15 \\ 
        \multicolumn{2}{l|}{POMDP} & 64 & 531.88 $\pm$ 158.97 & 722.45 $\pm$ 112.39 & 72 & 399.97 $\pm$ 113.49 & 1000.42 $\pm$ 250.99 & 48 & 591.21 $\pm$ 217.72 & 513.84 $\pm$ 154.45\\
        \midrule
        \rowcolor[gray]{0.9}
        \multicolumn{2}{l|}{\textbf{Ours: MILP}} & \textbf{100} & \textbf{113.72} $\pm$ 0.43 & 5.90 $\pm$ 1.46 & \textbf{100} & \textbf{112.98} $\pm$ 0.37 & 10.27 $\pm$ 1.94 & \textbf{84} & \textbf{196.02} $\pm$ 40.52 & 62.36 $\pm$ 4.07\\
        \bottomrule
    \end{tabular}
    \begin{tablenotes}[flushleft]
    \item SR, ET, and CT are success rate [\%], execution time [s], and computation time [s]. ET and CT: mean $\pm$ SD for successful executions across 25 runs at risk tolerance $\Delta_c = 0.1$. 
    \end{tablenotes}
   \end{threeparttable}
   \vspace{-4mm}
\end{table*}

\subsection{Results}
\label{sec:experiment_results}
\begin{figure}[t]
  \centering
  \begin{minipage}[b]{0.49\linewidth}
    \includegraphics[width=\linewidth]{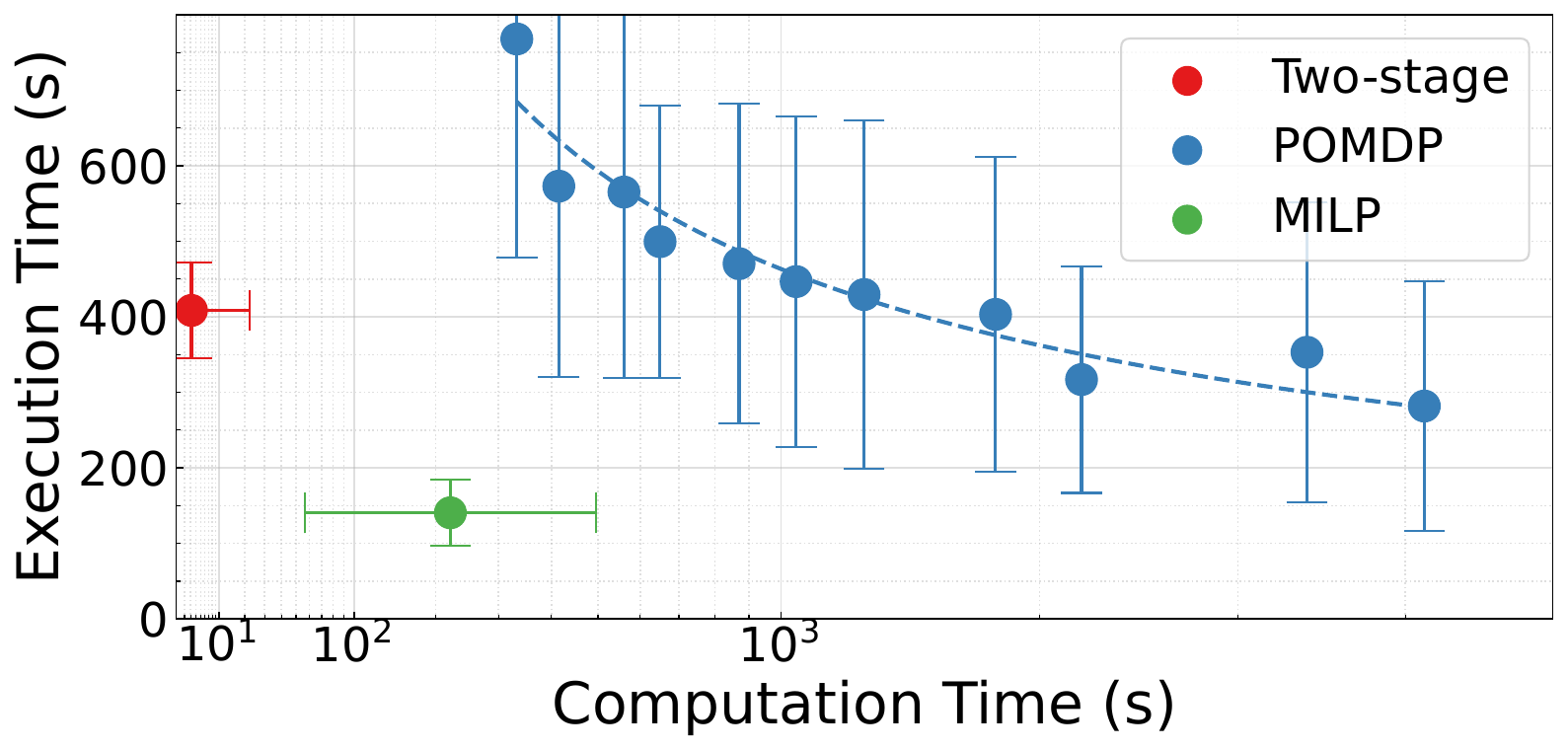}
    \subcaption{}
  \end{minipage}%
  \begin{minipage}[b]{0.49\linewidth}
    \includegraphics[width=\linewidth]{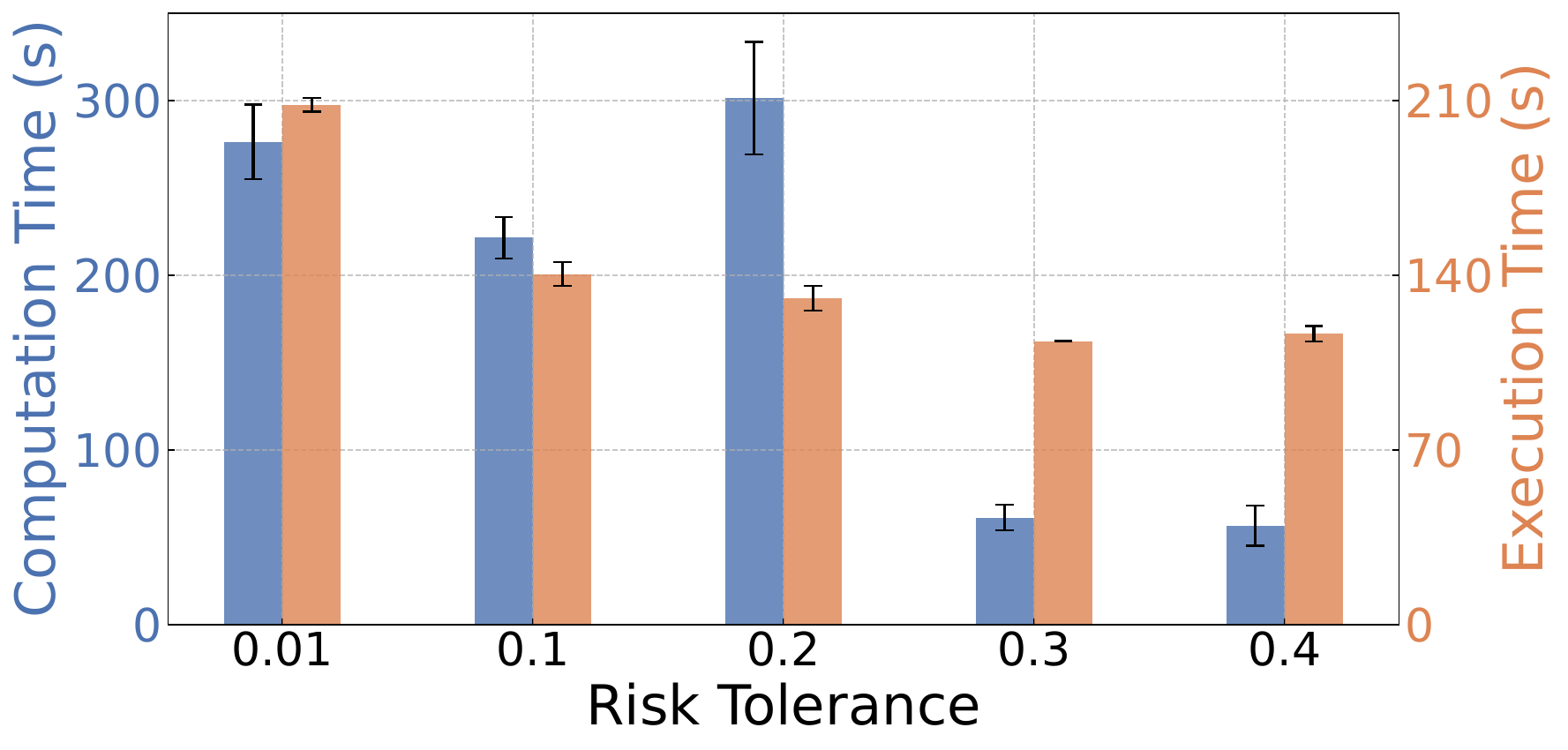}
    \subcaption{}
  \end{minipage}
  \vspace{-2mm}
  \caption{Results from ten random maps using MILP planner: (a) Comparison of CT and ET with other planners (x-axis in square root scale). (b) Effects of varying risk tolerances on scan frequency and planning computation time.}
  \vspace{-4mm}
  \label{fig:quantitative_plots}
\end{figure}

\emph{Quantitative Comparison:} 25 Monte Carlo simulations were run with $\Delta_c$ = 0.1 and Table \ref{tab:quantitative_comparison_three_configs} summarizes the mission planning performance across the three obstacle scenarios. 
Our approach achieves the best performance in navigation efficiency, as shown by the lowest ET.
This results from optimizing discrete actions for efficient navigation while minimizing costly scans and accounting for collision risks due to localization errors in a probabilistic manner.
The two-stage planner shows the lowest CT; however, its solution quality is worse than ours for navigation efficiency.
The POMDP planner struggles with TAMP under uncertainty, shown by poor planning outcomes and the highest CT.
This result indicates the exact solution of CC-HPOMDP is impractical due to computational intractability.
Fig. \ref{fig:quantitative_plots}a shows tradeoffs between CT and ET for ten Rnd-maps. The MILP planner achieves the lowest ET while maintaining low CT. The anytime POMDP planner improves over time but requires more computation than ours.

\begin{figure}[t]
  \centering
  \begin{minipage}[b]{0.333\linewidth}
    \includegraphics[width=\linewidth]{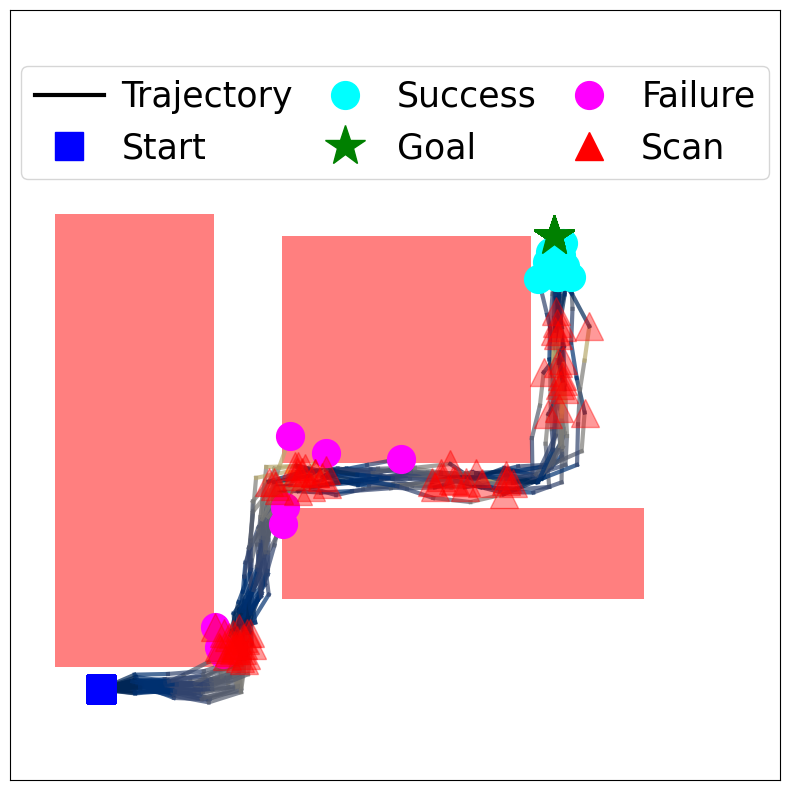}
    \subcaption{Two-stage}
  \end{minipage}%
  \begin{minipage}[b]{0.333\linewidth}
    \includegraphics[width=\linewidth]{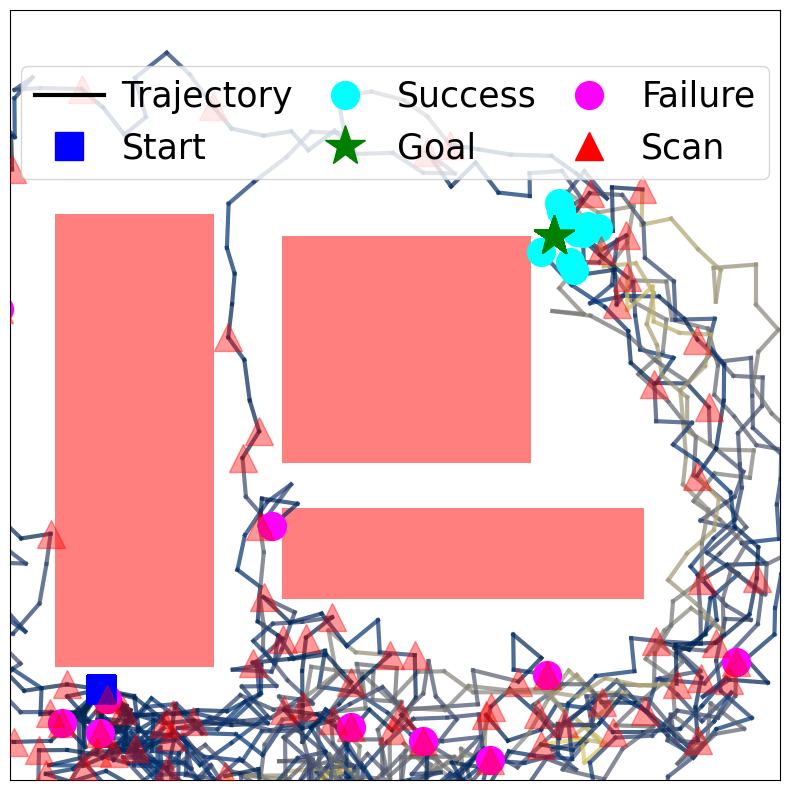}
    \subcaption{POMDP}
  \end{minipage}%
  \begin{minipage}[b]{0.333\linewidth}
    \includegraphics[width=\linewidth]{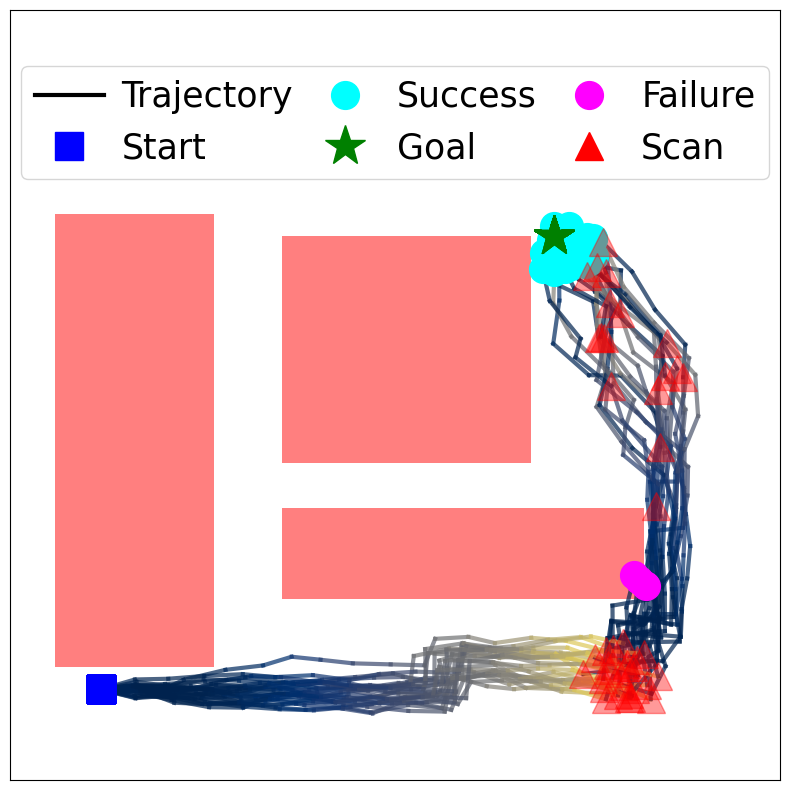}
    \subcaption{Ours: MILP}
  \end{minipage}
  \vspace{-6mm}
  \caption{Comparison of planning and execution in narrow map among three planners. Trajectories are colored by covariance growth during planning.}
  \vspace{-2mm}
  \label{fig:mc-trajectories-trap}
\end{figure}

Fig. \ref{fig:mc-trajectories-trap} shows Monte Carlo rollout results for Nar-map, color-coded by covariance growth in planning.
The two-stage planner often fails in narrow passages due to its decoupled approach: finding the shortest paths first, then scheduling perception along them.
The POMDP planner leads to stochastic execution due to 1) incomplete convergence in anytime planning, and 2) uncertainty sampling to address the curse of history in hyperbelief space.
Our MILP planner enables safe, efficient navigation: maintaining obstacle clearance during blind locomotion while scanning to reduce uncertainty.

\emph{Parameter Study:} Fig. \ref{fig:quantitative_plots}b shows the performance of our MILP planner with $\Delta_c \in \{0.01, \ldots, 0.4\}$ in ten Rnd-maps.
We confirm that integrating probability constraints into MILP is effective, as lower $\Delta_c$ values achieve more conservative executions.
Conversely, higher $\Delta_c$ leads to more frequent blind locomotion, allowing collision risks and reducing computation time by expanding the feasible solution space.

\begin{figure}[t]
     \centering
     \includegraphics[width=0.87\linewidth]{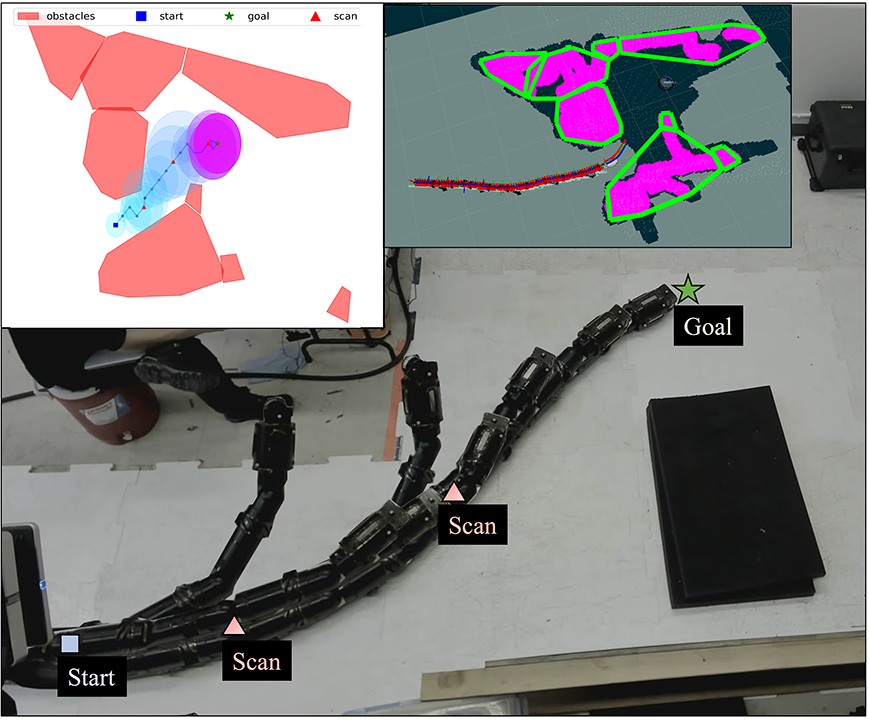}
     \caption{Navigation of EELS hardware using the MILP planner. It effectively schedules scans when the collision chance becomes high.}
     \vspace{-4mm}
     \label{fig:move-scan-hw}
 \end{figure}

\emph{Hardware Demonstration:} Our MILP planner has been integrated with the EELS platform for real-world testing, as shown in Fig. \ref{fig:move-scan-hw} presenting overlaid frames from a context camera. 
The top-left corner of the image displays a visualization of BLISS-TAMP's belief space planning, while the top-right corner shows the breakdown of obstacles into convex lines.
Our planner enables blind locomotion in open areas while effectively scheduling scans in narrow passages based on collision risk assessment for safe and efficient navigation.

\subsection{Limitations and Possible Extensions}
MILPs are NP-hard, with solution time scaling exponentially with the number of \textit{obstacles}, \textit{planning horizons}, and \textit{actions}.
To address these computational challenges, we will extend our current MILP formulation to a continuous-time approach, which would significantly reduce the optimization problem size and improve computational efficiency.

\section{Conclusion} 
\label{sec:conclusion}

We have introduced BLISS-TAMP, a resilient navigation strategy that alternates proprioceptive-only movement with exteroceptive scans for snake robots to achieve robustness against localization failures.
We formulate a novel chance-constrained convex MILP to approximate solutions for a partially observable TAMP problem, balancing movement and scanning to increase state knowledge while considering execution time.
Simulation studies validated that the proposed algorithm outperformed existing planners in terms of planning efficiency while reducing computational complexity.
Hardware experiments on EELS demonstrated real-world effectiveness, though quantitative evaluations are needed for field deployment.
Future work will focus on implementing continuous-time MILP to reduce computational complexity.

\bibliographystyle{IEEEtran}
\bibliography{references}

\newpage
\section{Supplemental Material}

\begin{figure*}[t]
  \centering
  \begin{minipage}[b]{0.166\linewidth}
    \includegraphics[width=\linewidth]{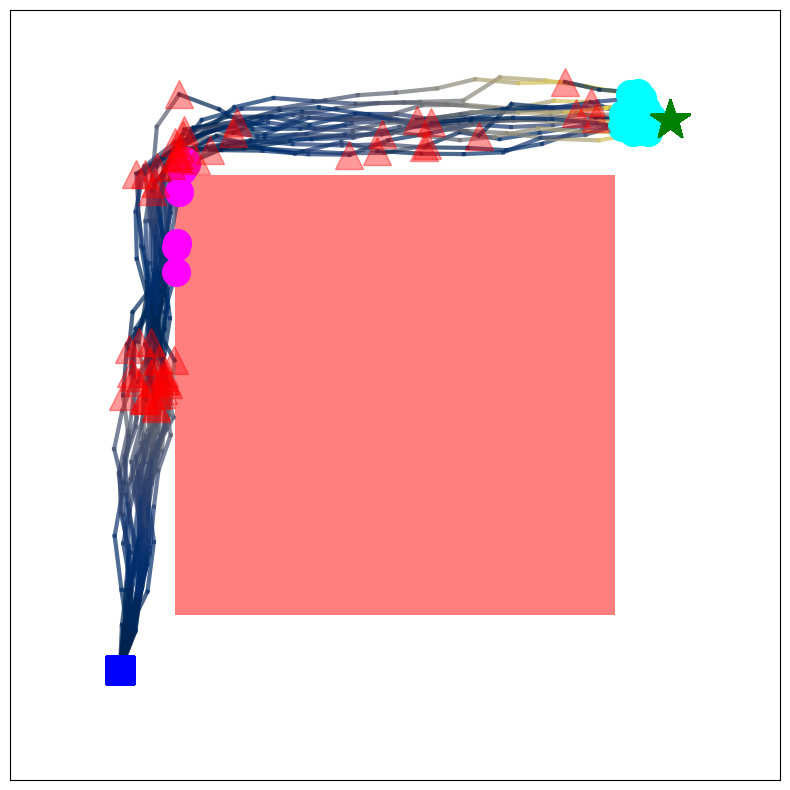}
    \subcaption{Two-stage}
  \end{minipage}%
  \begin{minipage}[b]{0.166\linewidth}
    \includegraphics[width=\linewidth]{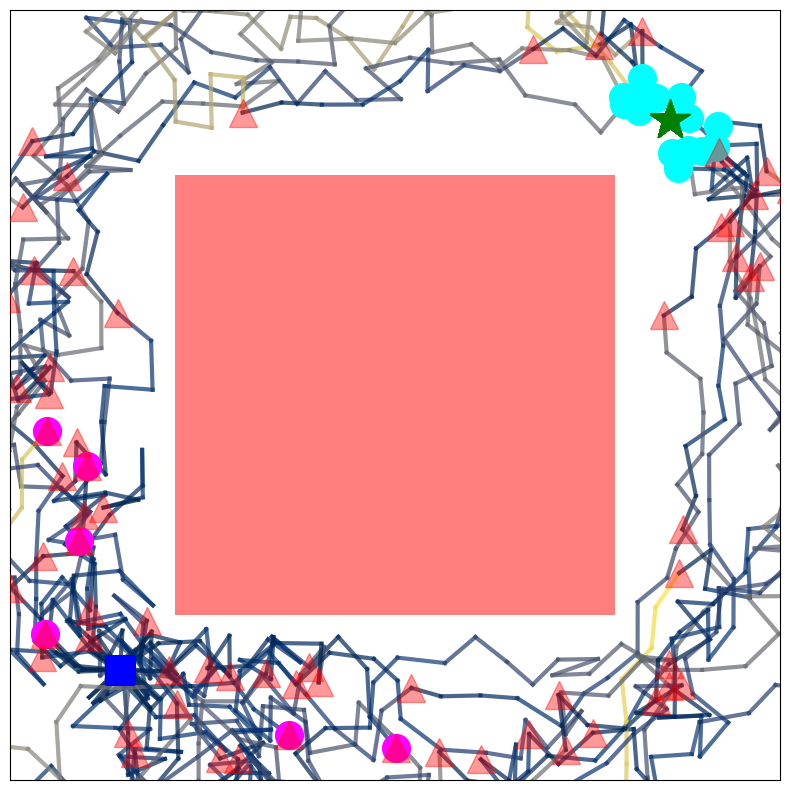}
    \subcaption{POMDP}
  \end{minipage}%
  \begin{minipage}[b]{0.166\linewidth}
    \includegraphics[width=\linewidth]{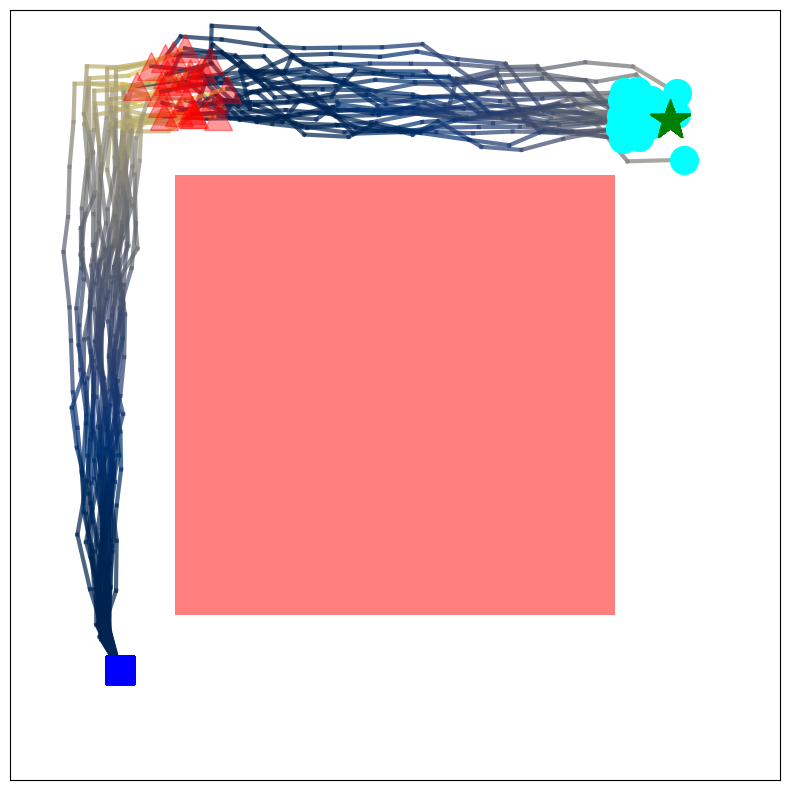}
    \subcaption{Ours: MILP}
  \end{minipage}%
  \begin{minipage}[b]{0.166\linewidth}
    \includegraphics[width=\linewidth]{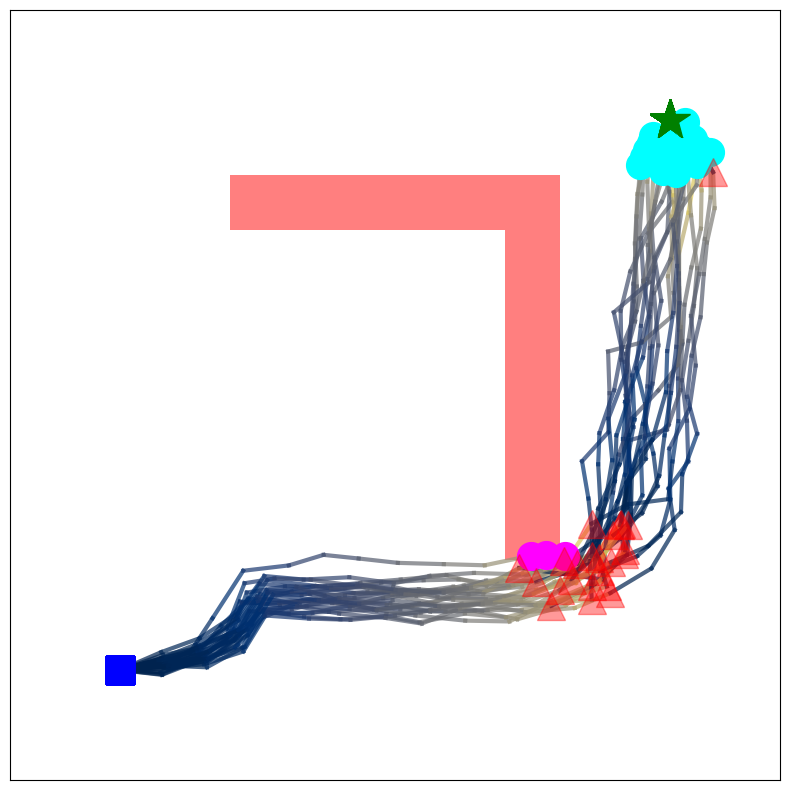}
    \subcaption{Two-stage}
  \end{minipage}%
  \begin{minipage}[b]{0.166\linewidth}
    \includegraphics[width=\linewidth]{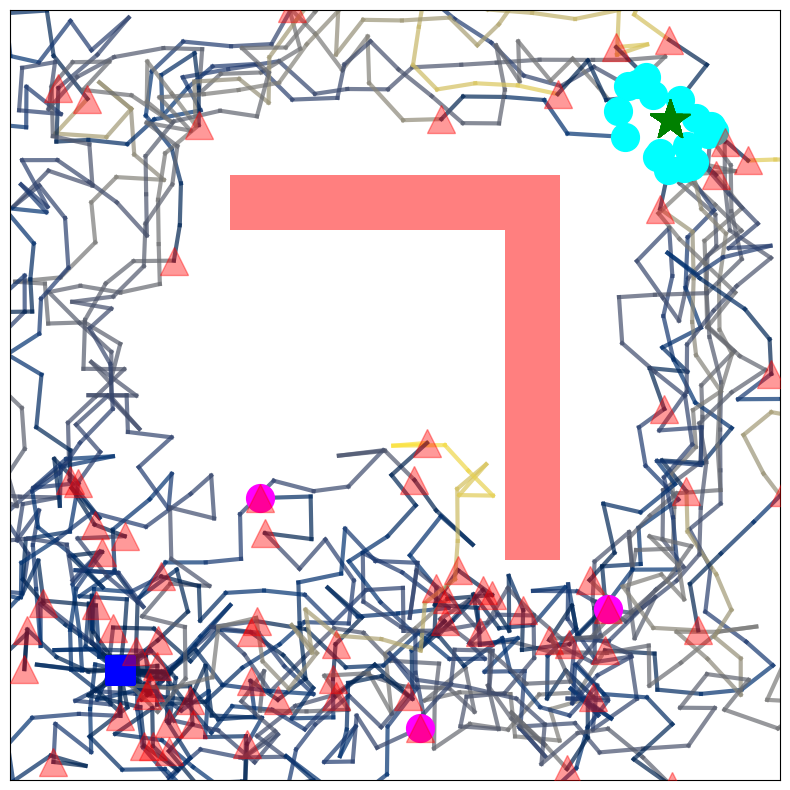}
    \subcaption{POMDP}
  \end{minipage}%
  \begin{minipage}[b]{0.166\linewidth}
    \includegraphics[width=\linewidth]{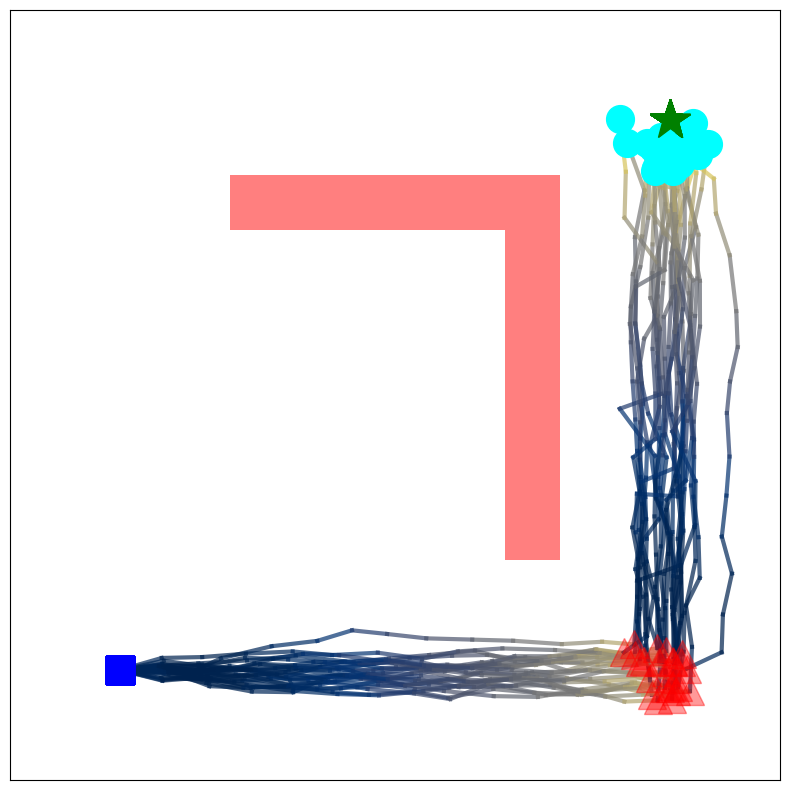}
    \subcaption{Ours: MILP}
  \end{minipage}%
\caption{Comparison of planning and execution among three planners. Markers and colors of trajectories represent the same meanings as in Fig. \ref{fig:mc-trajectories-trap}. (a)-(c) show results in the standard map, and (d)-(f) in the entrapped map. The MILP planner enables conservative navigation to avoid collision risks while effectively scheduling scans based on uncertainty growth. The two-stage planner prioritizes path distance optimality, which results in obstacle collisions. The POMDP planner produces highly stochastic trajectories due to its anytime nature and the inherent difficulty in solving this problem exactly.}
  \vspace{-2mm}
  \label{fig:comparison_rect_trap}
\end{figure*}

\begin{figure*}[t]
\captionsetup[sub]{justification=centering}
  \centering
  \begin{minipage}[b]{0.199\linewidth}
    \includegraphics[width=\linewidth]{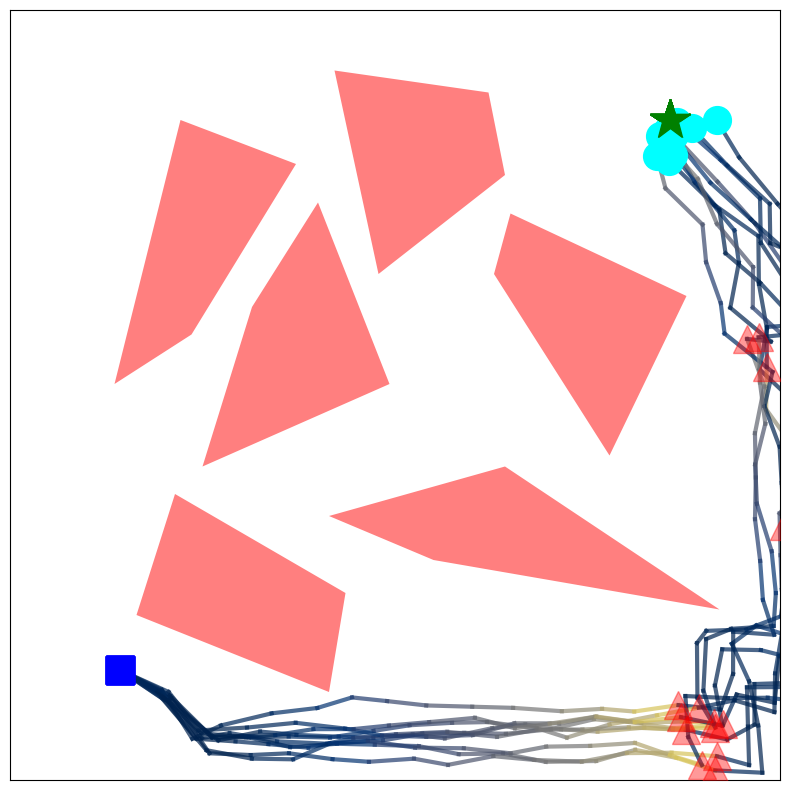}
    \subcaption{$\Delta_c=0.01$\\
                Execution Time: 215.80\\
                Computation Time: 812.06}
  \end{minipage}%
  \begin{minipage}[b]{0.199\linewidth}
    \includegraphics[width=\linewidth]{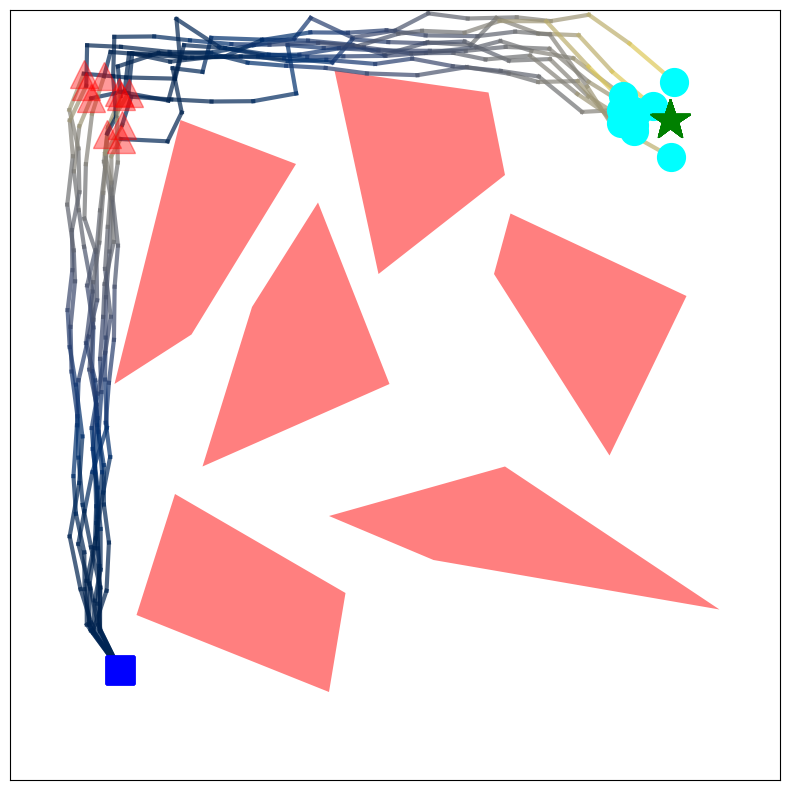}
    \subcaption{$\Delta_c=0.1$\\
                Execution Time: 114.35\\
                Computation Time: 307.82}
  \end{minipage}%
  \begin{minipage}[b]{0.199\linewidth}
    \includegraphics[width=\linewidth]{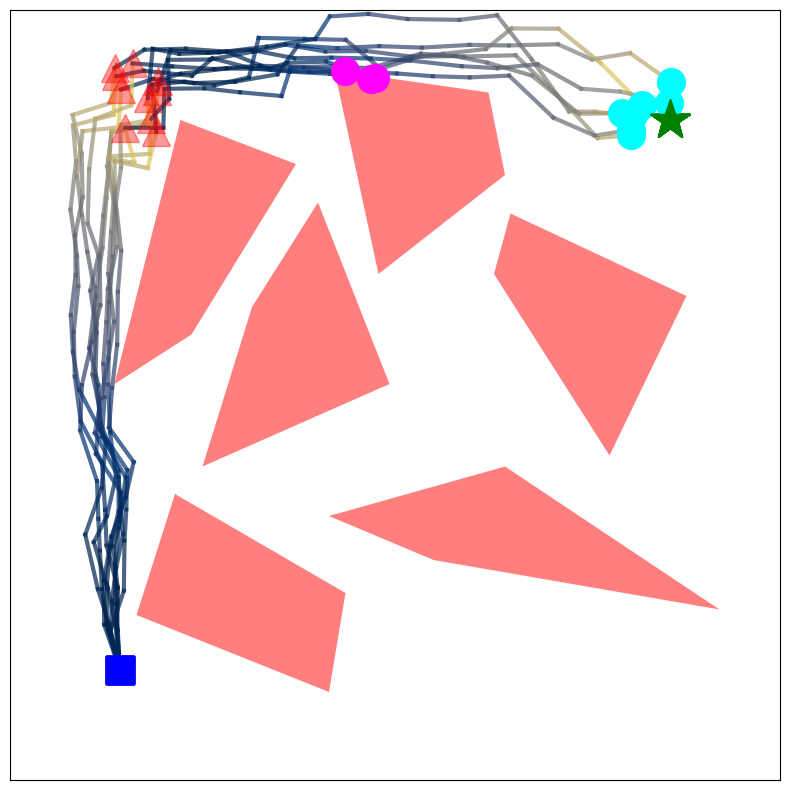}
    \subcaption{$\Delta_c=0.2$\\
                Execution Time: 114.29\\
                Computation Time: 404.42}
  \end{minipage}%
  \begin{minipage}[b]{0.199\linewidth}
    \includegraphics[width=\linewidth]{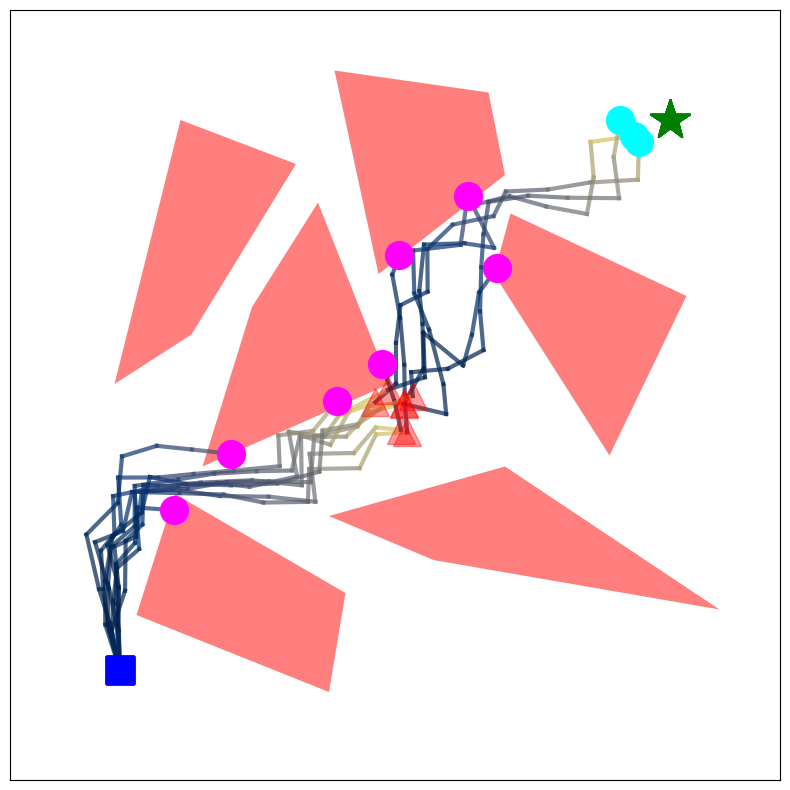}
    \subcaption{$\Delta_c=0.3$\\
                Execution Time: 112.67\\
                Computation Time: 24.54}
  \end{minipage}%
  \begin{minipage}[b]{0.199\linewidth}
    \includegraphics[width=\linewidth]{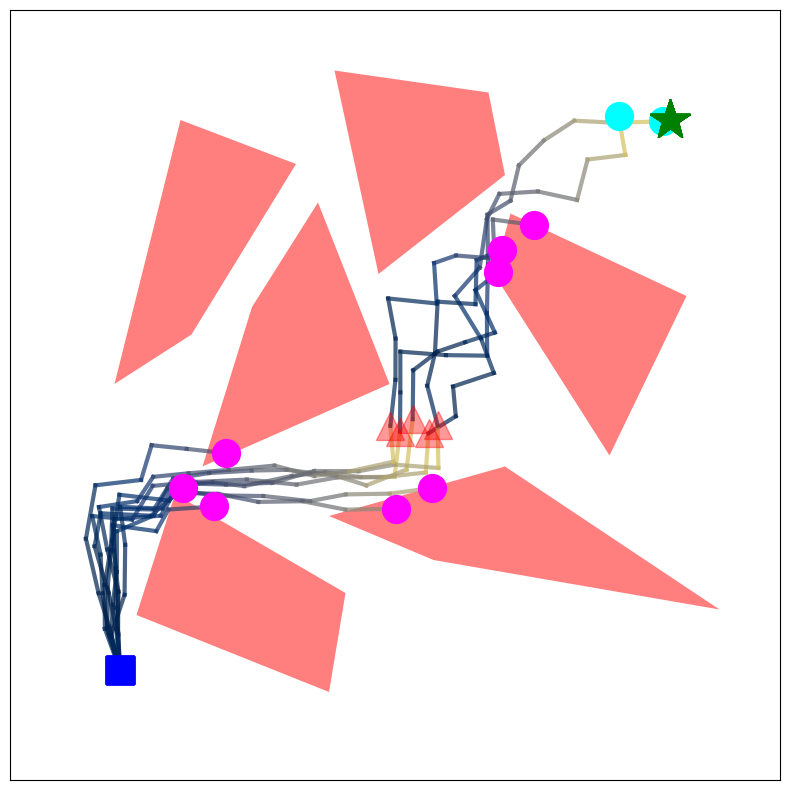}
    \subcaption{$\Delta_c=0.4$\\
                Execution Time: 113.00\\
                Computation Time: 21.30}
  \end{minipage}%
    \caption{Comparison of MILP's planning and execution on a random map with different risk tolerances with execution and computation times. Markers and trajectories' colors represent the same meaning as in Fig. \ref{fig:mc-trajectories-trap}. As $\Delta_c$ increases from lower to higher values, navigation transitions from conservative to aggressive, allowing entry into narrow passages to minimize path lengths while also increasing collision chances.}
  \vspace{-2mm}
  \label{fig:milp_trajectories_different_risk}
\end{figure*}

This supplement provides simulation visualizations and details how risk tolerance affects planning behaviors.

\subsection{Qualitative Comparison against Comparable Planners}

Fig. \ref{fig:comparison_rect_trap} presents the remaining results of Monte Carlo simulations with $\Delta_c=0.1$ for Ent- and Nar-maps, comparing the MILP planner with two other planners in Section \ref{sec:experiment_results}.
The MILP planner maintains obstacle clearance for safe blind locomotion while scheduling scans to prevent excessive localization uncertainty growth (Fig. \ref{fig:comparison_rect_trap}c,f). 
This strategy balances safety from collisions and efficiency in execution time, accomplished by simultaneously addressing task planning for scanning and motion planning for navigation.
A decoupled solution cannot attain this balance, as it prioritizes preprocessed decision-making.
This can be observed in the two-stage planner, which often fails to avoid collisions despite planning frequent scans (Fig. \ref{fig:comparison_rect_trap}a,d).

It is also important that our MILP planner achieves significantly higher computational efficiency and more consistent TAMP compared to the POMDP planner.
This advantage stems from our use of convex deterministic optimization, which eliminates the need for uncertainty samples through the belief space assumption and simplifies the problem formulation.
Otherwise, BLISS-TAMP requires exact solutions to CC-HPOMDP and handles uncertainty through sampling, which causes computationally demanding and inconsistent planning outcomes (Fig. \ref{fig:comparison_rect_trap}b,e).

\subsection{Qualitative Analysis of Risk Tolerance Effects on MILP}

Fig. \ref{fig:milp_trajectories_different_risk} shows Monte Carlo simulation results for our MILP planner with $\Delta_c=\{0.01, \ldots, 0.4\}$ in one of ten Rnd-maps used for parameter study.
The given trajectories and associated scans indicate that our integration of chance constraints into MILP functions as improving safety.
Lower $\Delta_c$ provides higher obstacle clearance and more frequent scans to reduce uncertainty growth, which are important to avoid collisions under uncertainty.
As we increase $\Delta_c$, the MILP planner allows more aggressive maneuvering, such as passing through narrow passages (Fig. \ref{fig:comparison_rect_trap}d,e).
As expressed in reduced ET, higher $\Delta_c$ contributes to more distance-efficient mission execution but also requires taking risks in uncertain scenarios.
Our method offers the ability to adjust planning behavior through simple parameter tuning.
This controllability is beneficial as the balance between safety and efficiency changes throughout a mission's timeline.
For example, during the initial exploration phase when most terrain is unknown, it may be desirable to prioritize safety for long-term mission resilience.

\end{document}